\documentclass[journal]{vgtc}                     % final (journal style)
\onlineid{0}

\vgtccategory{Research}

\vgtcpapertype{algorithm/technique}

\title{Spatiotemporal Calibration and Ground Truth Estimation for High-Precision SLAM Benchmarking in Extended Reality}

\author{%
  \authororcid{Zichao Shu}{0009-0000-4785-6234},
  \authororcid{Shitao Bei}{0009-0003-0601-5414},
  \authororcid{Lijun Li}{0009-0007-0818-6929}, and 
  \authororcid{Zetao Chen}{0000-0002-5596-5008}
}

\authorfooter{
  \item
        Zichao Shu is with the Advanced Display and Sensing Research Center, Yongjiang Laboratory, Ningbo, China, and also with the Shanghai Institute of Technical Physics, Chinese Academy of Sciences, Shanghai, China. 
  	E-mail: zichao-shu@ylab.ac.cn
  \item
        Shitao Bei, Lijun Li, and Zetao Chen are with the Advanced Display and Sensing Research Center, Yongjiang Laboratory, Ningbo, China. 
  	E-mail: \{shitao-bei, lijun-li, zetao-chen\}@ylab.ac.cn
  \item Corresponding Author: Zetao Chen
}

\abstract{
  Simultaneous localization and mapping (SLAM) plays a fundamental role in extended reality (XR) applications. As the standards for immersion in XR continue to increase, the demands for SLAM benchmarking have become more stringent. Trajectory accuracy is the key metric, and marker-based optical motion capture (MoCap) systems are widely used to generate ground truth (GT) because of their drift-free and relatively accurate measurements. However, the precision of MoCap-based GT is limited by two factors: the spatiotemporal calibration with the device under test (DUT) and the inherent jitter in the MoCap measurements. These limitations hinder accurate SLAM benchmarking, particularly for key metrics like rotation error and inter-frame jitter, which are critical for immersive XR experiences. This paper presents a novel continuous-time maximum likelihood estimator to address these challenges. The proposed method integrates auxiliary inertial measurement unit (IMU) data to compensate for MoCap jitter. Additionally, a variable time synchronization method and a pose residual based on screw congruence constraints are proposed, enabling precise spatiotemporal calibration across multiple sensors and the DUT. Experimental results demonstrate that our approach outperforms existing methods, achieving the precision necessary for comprehensive benchmarking of state-of-the-art SLAM algorithms in XR applications. Furthermore, we thoroughly validate the practicality of our method by benchmarking several leading XR devices and open-source SLAM algorithms. The code is publicly available at \url{https://github.com/ylab-xrpg/xr-hpgt}.
}


\teaser{
  \centering
  \includegraphics[width=0.92\linewidth]{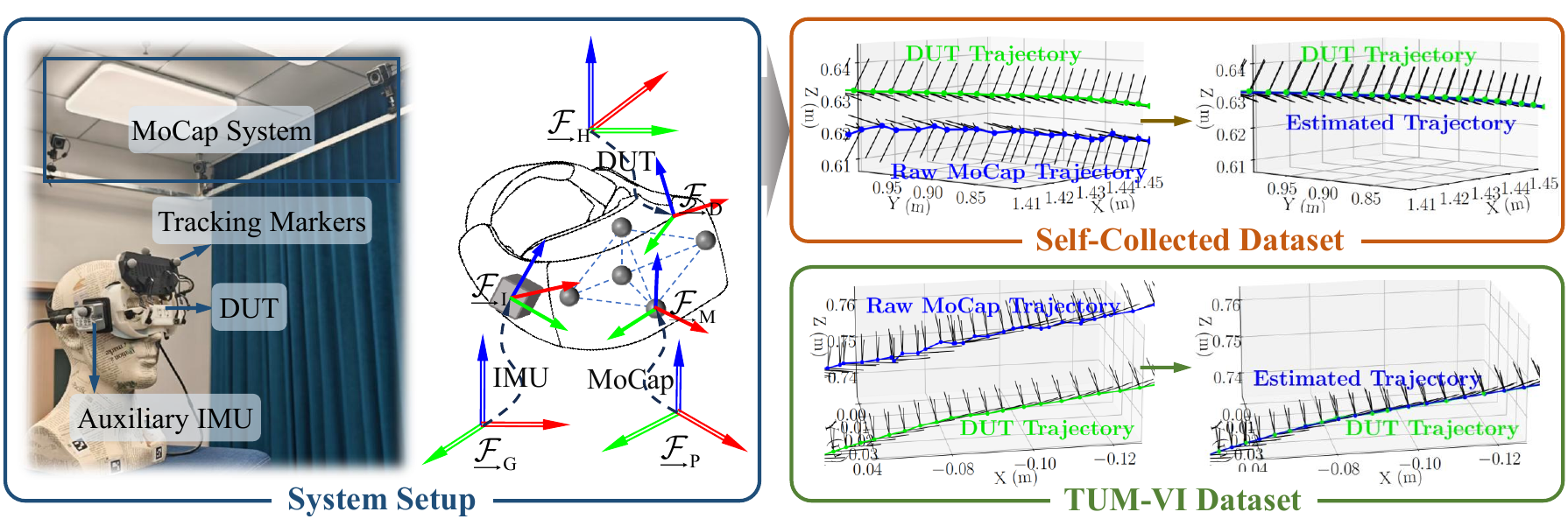}
  \caption{
    We propose a localization GT estimator for high-precision benchmarking of XR-oriented SLAM. The left panel shows the sensor setup, and the right presents estimated trajectories on self-collected and public datasets. It greatly mitigates jitter in the raw MoCap data (amplified for visualization) and achieves precise spatiotemporal calibration across multiple sensors and the DUT. The resulting trajectories are smoothly refined and well-aligned, serving as a reliable benchmark for evaluating the SLAM performance of the DUT.
  }
  \label{FIG: Teaser figure}
}

\graphicspath{{figs/}{figures/}{pictures/}{images/}{./}} % where to search for the images

\usepackage{amsmath}
\usepackage{multirow}
\usepackage{booktabs}
\usepackage{bm}
\usepackage{graphicx}
\usepackage{array}
\usepackage{makecell}
\usepackage{adjustbox}
\usepackage{xcolor}
\usepackage{pifont}
\usepackage{colortbl}
\usepackage{amsfonts}
\usepackage{mathptmx}                  % use matching math font

\definecolor{myred}{RGB}{200,0,0}
\newcommand{\xmark}{\textcolor{myred}{\ding{55}}}

\definecolor{mygreen}{RGB}{0,128,0}
\newcommand{\cmark}{\textcolor{mygreen}{\ding{51}}}

\DeclareMathAlphabet{\mathcal}{OMS}{cmsy}{m}{n}

\setlist[itemize]{itemsep=1pt, parsep=2pt, topsep=3pt}

\begin{document}

%%%%%%%%%%%%%%%%%%%%%%%%%%%%%%%%%%%%%%%%%%%%%%%%%%%%%%%%%%%%%%%%
%%%%%%%%%%%%%%%%%%%%%% START OF THE PAPER %%%%%%%%%%%%%%%%%%%%%%
%%%%%%%%%%%%%%%%%%%%%%%%%%%%%%%%%%%%%%%%%%%%%%%%%%%%%%%%%%%%%%%%

\maketitle

\section{Introduction} \label{SEC: Introduction}

Extended reality (XR), an umbrella term for virtual reality (VR), augmented reality (AR) and mixed reality (MR), has emerged as a promising next-generation computing platform. A key feature of XR is displaying virtual content that remains stationary relative to the physical world as users move, known as world-locked (WL) rendering \cite{guan2023perceptual}. This capability facilitates the extension of physical reality, ensuring consistent virtual-real alignment for immersive experiences. Accurate six-degree-of-freedom (6-DoF) head pose estimation is crucial for WL rendering and is typically achieved through visual-inertial simultaneous localization and mapping (VI-SLAM) in modern XR devices \cite{tran2023wearable, pan2024robust, miyake2024development}. However, inherent limitations in sensor precision and algorithmic performance inevitably introduce pose errors, which primarily manifest as trajectory drift and jitter. These errors lead to imperfect WL rendering that causes perceptual artifacts and degrades the user experience. Thus, systematically benchmarking SLAM trajectories is essential for the development and optimization of XR systems. \cite{sheng2024review}.

SLAM benchmarking typically involves comparing the trajectory estimated by the onboard SLAM algorithm of the device under test (DUT) with high-precision ground truth (GT), followed by the computation of error metrics. Among these, commonly used metrics include absolute rotation/translation error (ARE/ATE) to quantify trajectory drift and relative rotation/translation error (RRE/RTE) to measure localization jitter \cite{sturm2012benchmark, jinyu2019survey}. In near-eye XR devices, such as head-mounted displays (HMDs) and glasses-style devices, even small motion errors can significantly impact the user experience \cite{burr2011motion, wilmott2022perceptibility, lutwak2023user}. State-of-the-art (SOTA) XR SLAM algorithms can achieve ARE/ATE below 1°/10 mm and RRE/RTE below 0.1°/1 mm in typical indoor scenarios \cite{campos2021orb, rosinol2020kimera, geneva2020openvins, qin2018vins}, enabling high-immersion WL rendering. As SLAM accuracy improves, obtaining GT of greater precision becomes increasingly challenging. To ensure accurate quantification, the GT error should be smaller than that of SLAM by at least half an order of magnitude. In existing works, marker-based optical motion capture (MoCap) systems, such as Vicon or OptiTrack, have commonly been used to provide indoor localization GT \cite{swamy2024appear, banaszczyk2024accurate, cheng2024comparing, dong2024sevar}. These systems track infrared reflective markers attached to the DUT, providing drift-free 6-DoF pose estimates through outside-in localization \cite{liu2024mvins, monica2024adaptive}.

However, MoCap accuracy is limited by two main factors. First, the MoCap system and the DUT operate with independent clocks and reference frames, requiring spatiotemporal calibration. In practice, the lack of communication between the two systems often causes clock scale drift, which can exceed 2 ms per minute for consumer-grade devices \cite{mueggler2017event}. Additionally, commercial DUTs typically do not provide access to raw sensor data, which necessitates performing calibration based on output trajectories that inherently contain cumulative errors. These issues can result in an ARE above 0.3° and an ATE greater than 2 mm \cite{shu2024spatiotemporal, furrer2018evaluation, zhou2023simultaneously}. Second, although MoCap systems generally offer high absolute accuracy, their vision-based marker tracking is susceptible to unstable imaging and segmentation. This introduces high-frequency jitter at the system output rate, resulting in inter-frame RRE and RTE typically exceeding 0.15° and 0.5 mm, respectively \cite{burri2016euroc, schubert2018tum}. Visualization results can be seen on the right side of \cref{FIG: Teaser figure}. These errors, combined with calibration inaccuracies, further reduce GT precision. While some studies attempt to suppress jitter by fusing MoCap data with DUT IMU measurements \cite{geneva2020vicon2gt, burri2016euroc}, the low accuracy and instability of consumer-grade IMUs limit their effectiveness. In summary, although current MoCap-based GT estimation methods provide satisfactory absolute translation accuracy, they are insufficient for evaluating rotation error and inter-frame jitter in SOTA SLAM algorithms. These metrics are often overlooked in other SLAM domains but are crucial for user experience in XR applications \cite{atac2022motion, wilmott2022perceptibility, lutwak2023user}.

This work aims to develop a MoCap-based localization GT estimation method. To ensure broad applicability and enable benchmarking of commercial devices, the method operates directly on DUT output poses without requiring raw sensor data. It is designed to meet the aforementioned accuracy requirements for SLAM benchmarking, targeting ARE/ATE $<$ 0.2°/2 mm and RRE/RTE $<$ 0.02°/0.2 mm. To this end, we propose a novel method that introduces an auxiliary IMU and integrates MoCap, IMU, and DUT data in a maximum likelihood estimation (MLE) framework, enabling efficient sensor calibration and MoCap jitter compensation. To the best of our knowledge, this is the first method to meet all the above targets (see \cref{TAB: Trajectory estimation accuracy} later for details). The main contributions of this work are summarized as follows:

\begin{itemize}
\item We propose a continuous-time state estimator that performs joint spatiotemporal calibration across all involved sensors. By exploiting the complementary strengths of different inputs in both the spatial and derivative domains, the method effectively balances errors and produces high-precision localization GT for SLAM benchmarking.
\item To correct time drift between independent clocks and reduce cumulative errors in DUT data, we propose a novel variable time offset model and a DUT pose factor based on screw congruence constraints to enhance our estimator.
\item Extensive experiments on both simulated and real-world datasets demonstrate that our approach outperforms SOTA MoCap-based GT estimation methods. The code is open-sourced to benefit the community.
\item Based on our estimator, we benchmark multiple leading XR devices and SLAM algorithms. Unlike previous evaluations limited by GT inaccuracies, our approach is the first to enable comprehensive and precise error quantification, particularly for rotational drift and inter-frame jitter, providing a reliable reference for the advancement of SLAM in the XR field.
\end{itemize}

\section{Related Works} \label{SEC: Related works}

To achieve high-precision GT trajectory estimation, we focus on two key challenges: spatiotemporal calibration and MoCap jitter mitigation. Furthermore, continuous-time batch estimation can potentially address both. We analyze each aspect and discuss limitations in existing methods.

\subsection{Spatiotemporal Calibration}

For most studies on MoCap-based localization GT estimation, the focus has generally been limited to spatiotemporal calibration between the MoCap system and the DUT \cite{gao2022vector, lee2022vivid++, zhu2018multivehicle}. This is typically modeled as a hand-eye calibration problem, where the MoCap system is treated as the `hand' and the DUT as the `eye'. The process involves time synchronization followed by estimating the homogeneous transformation between their body frames \cite{wu2020correspondence}.

For time synchronization, the problem can be formulated as a motion registration task, which can be solved using iterative closest point algorithms \cite{kelly2014general}. Alternatively, a more common approach is based on cross-correlation theory \cite{li2015simultaneous, pachtrachai2018chess, ackerman2013sensor}, where the time offset is determined by maximizing the cross-correlation between discrete motion signals from the collected data. For spatial calibration, the methodology depends on the type of data provided by the DUT. When the raw sensor data is accessible, precise calibration can be achieved by integrating MoCap measurements with the calibration pattern information captured by the DUT's camera. This allows for independent estimation of the camera poses, which can then be solved by formulating it as the $\bm{AX}=\bm{XB}$ problem \cite{zhou2023simultaneously, wu2021simultaneous, sarabandi2022hand}. Alternatively, tightly coupled optimization can be applied, where reprojection constraints of the calibration pattern corners are used to determine the results \cite{pedrosa2021general, koide2019general, zhong2023robot, jinyu2019survey}. In more general cases, such as commercial XR devices, raw sensor data is often unavailable. As a result, calibration must directly rely on the trajectories provided by the DUT. Since these measurements inherently contain cumulative errors, some studies propose constructing constraints using relative poses derived from trajectory segments \cite{shu2024spatiotemporal, furrer2018evaluation}. A limitation of these methods is their assumption of equal weights for all residuals. A more robust approach involves estimating the variance for each measurement to guide the solution.

\subsection{Jitter Mitigation}

MoCap jitter mitigation has received limited attention.  In conventional SLAM applications or handheld XR devices like smartphones, jitter has minimal impact on performance, so benchmarking typically focuses on absolute errors such as ATE \cite{li2024rd, zhang2024100, leutenegger2015keyframe}. However, in highly immersive near-eye XR devices, jitter tolerance is much lower \cite{wilmott2022perceptibility, lutwak2023user}. Excessive jitter not only degrades user experience but also contributes to action difficulties \cite{ham2021we} and simulator sickness \cite{stauffert2018effects}. Thus, mitigating MoCap jitter becomes essential to enable accurate quantification of inter-frame RRE and RTE.

Some studies have addressed this problem by tightly coupling MoCap and IMU data within the MLE framework, reducing jitter while enabling the simultaneous spatiotemporal calibration. For instance, the EuRoC MAV dataset utilized a batch estimator to fuse MoCap and IMU measurements, obtaining the GT trajectory \cite{burri2016euroc}. The Vicon2GT toolbox, which integrates IMU preintegration and MoCap pose constraints into a joint optimization problem to refine the GT trajectory \cite{geneva2020vicon2gt}, and it has been adopted for benchmarking several SLAM algorithms \cite{chen2023monocular, katragadda2024nerf, zhang2022dido}. As a further improvement, MoCap2GT models MoCap measurements as interpolations on the $SE\left(3\right)$ manifold, and incorporates degradation detection to improve trajectory estimation accuracy \cite{shu2025mocap2gt}. These methods pioneered the fusion of MoCap and IMU data but did not explicitly explain the underlying mechanisms for jitter mitigation, which we will discuss in \cref{SEC: IMU Factors}. Additionally, they rely on the built-in IMU of the DUT, which lacks flexibility and may introduce uncontrollable errors. A more ideal solution would involve using an additional high-precision IMU. However, this also necessitates additional spatiotemporal calibration, and the methods mentioned above do not support such multi-sensor configurations, highlighting the need for a more general framework. Moreover, when estimating high-frequency GT trajectories, preintegration-based approaches are prone to degeneracy \cite{forster2016manifold, eckenhoff2019closed}, which further restricts their effectiveness. On the other hand, some alternative approaches focus on directly smoothing the poses output by MoCap systems. For example, the TUM-VI dataset applies median filtering to reduce jitter \cite{schubert2018tum}. Although this method mitigates jitter to some extent, it compromises the precision of MoCap trajectories.

\subsection{Continuous-Time MLE}

A continuous-time MLE framework provides an efficient solution for fusing data from MoCap, auxiliary IMU, and DUT, enabling concurrent sensor calibration and GT trajectory estimation. In this context, the rigid body motion is parameterized as a continuous-time function \cite{rehder2016general, furgale2012continuous}, such as B-splines on manifolds \cite{sommer2020efficient}, which makes it particularly well-suited to this application. This approach offers two key advantages. First, it enables flexible and consistent temporal sampling, allowing the general fusion of high-frequency data without degradation \cite{chen2024ikalibr}. Second, its inherent time differentiability facilitates the temporal calibration of different sensors without extra modeling \cite{cioffi2022continuous}, reducing potential sources of error. Continuous-time optimization has already been successfully applied to multi-sensor calibration tasks involving cameras, IMUs, and LiDARs \cite{chen2025ikalibr, li2024targetless, li2023two}. Because of its strong generalizability, it is also compatible with MoCap and DUT data. For example, the well-known Kalibr \cite{rehder2016extending} calibration framework has been extended in its MoCap branch to achieve this objective.

\begin{table}[]
\caption{Comparison of MoCap-based localization GT estimation methods among representative related works. Unlike previous methods, our approach fuses high-frequency data within a continuous-time MLE framework, and handles variable time offsets and trajectory drift. Here, HEC denotes hand-eye calibration.}
\label{TAB: Comparison of MoCap-based localization GT estimation methods}
\centering 
\footnotesize
\setlength{\tabcolsep}{5pt}
\begin{tabular}{c|ccccc}
\toprule[1.0pt]
\rotatebox{90}{\makecell[l]{Dataset/ \\ Method}} & \rotatebox{90}{\makecell[l]{Spatiotemporal \\ Calibration}} & \rotatebox{90}{\makecell[l]{Jitter \\ Mitigation}} & \rotatebox{90}{\makecell[l]{Continuous\\-Time MLE}} & \rotatebox{90}{\makecell[l]{Variable \\ Time Offset}} & \rotatebox{90}{\makecell[l]{Handle DUT \\ Trajectory Drift}} \\ \midrule 
\rule{0pt}{8pt} EuRoC \cite{burri2016euroc} & \multicolumn{2}{c}{Batch Estimation} & \multicolumn{1}{c}{\xmark} & \multicolumn{1}{c}{\xmark} & \multicolumn{1}{c}{\xmark} \\
\rule{0pt}{8pt} TUM-VI \cite{schubert2018tum} & \makecell{HEC} & \makecell{Median Filter} & \multicolumn{1}{c}{\xmark} & \multicolumn{1}{c}{\xmark} & \multicolumn{1}{c}{\xmark} \\
\rule{0pt}{8pt} Li et al. \cite{jinyu2019survey} & \makecell{HEC}  & \multicolumn{1}{c}{\xmark} & \multicolumn{1}{c}{\xmark} & \multicolumn{1}{c}{\xmark} & \multicolumn{1}{c}{\xmark} \\ 
\rule{0pt}{8pt} Vicon2GT \cite{geneva2020vicon2gt} & \multicolumn{2}{c}{Batch Estimation} & \multicolumn{1}{c}{\xmark} & \multicolumn{1}{c}{\xmark} & \multicolumn{1}{c}{\xmark} \\
\rule{0pt}{8pt} Kalibr-MoCap \cite{rehder2016extending} & \multicolumn{2}{c}{Batch Estimation} & \multicolumn{1}{c}{\cmark} & \multicolumn{1}{c}{\xmark} & \multicolumn{1}{c}{\xmark} \\  
\rule{0pt}{8pt} Shu et al. \cite{shu2024spatiotemporal} & \makecell{HEC}  & \multicolumn{1}{c}{\xmark} & \multicolumn{1}{c}{\cmark} & \multicolumn{1}{c}{\xmark} & \multicolumn{1}{c}{\cmark} \\ \midrule 
\rule{0pt}{8pt} Ours  & \multicolumn{2}{c}{Batch Estimation} & \multicolumn{1}{c}{\cmark} & \multicolumn{1}{c}{\cmark} & \multicolumn{1}{c}{\cmark} \\
\bottomrule[1.0pt]
\end{tabular}
\vspace{-8pt}
\end{table}

However, existing methods face several critical challenges. First, traditional calibration approaches often assume a fixed time offset between sensors, relying on inter-sensor communication or shared triggering mechanisms \cite{huai2022continuous, fu2021high, wang2024hardware}. In XR benchmarking scenarios, commercial devices typically operate on independent clocks, leading to unavoidable clock drift and time-varying offsets. For consumer-grade devices, this drift can exceed 2 ms per minute \cite{mueggler2017event}, which becomes non-negligible for high-precision GT estimation, requiring explicit modeling and estimation of the variable time offset. Second, since calibration is based on the DUT trajectory, which inherently contains accumulated errors \cite{shu2024spatiotemporal}, it is crucial to design a measurement factor that mitigates this issue. As a summary, \cref{TAB: Comparison of MoCap-based localization GT estimation methods} compares representative MoCap-based localization GT estimators, and our method is specifically designed to address the aforementioned challenges.

\section{Notation and Overview} \label{SEC: Notation and Overview}

In this section, we define the notation employed throughout the paper and give an overview of the problems we seek to address. The conventions for the coordinate frames are illustrated in the system setup on the left side of \cref{FIG: Teaser figure}.

\subsection{Notation}

\begin{itemize}
\item $\underrightarrow{\mathcal{F}}_{\scriptscriptstyle\mathrm{P}}, \underrightarrow{\mathcal{F}}_{\scriptscriptstyle\mathrm{M}}$: World and body coordinate frames of the MoCap system, where $\underrightarrow{\mathcal{F}}_{\scriptscriptstyle\mathrm{M}}$ also defines the clock of the MoCap.
\item $\underrightarrow{\mathcal{F}}_{\scriptscriptstyle\mathrm{G}}, \underrightarrow{\mathcal{F}}_{\scriptscriptstyle\mathrm{I}}$: World and body frames of the auxiliary IMU, where $\underrightarrow{\mathcal{F}}_{\scriptscriptstyle\mathrm{I}}$ also defines the clock of the IMU.
\item $\underrightarrow{\mathcal{F}}_{\scriptscriptstyle\mathrm{H}}, \underrightarrow{\mathcal{F}}_{\scriptscriptstyle\mathrm{D}}$: World and body frames of the DUT, where $\underrightarrow{\mathcal{F}}_{\scriptscriptstyle\mathrm{D}}$ also defines the clock of the DUT.
\item $\underrightarrow{\mathcal{F}}_{\scriptscriptstyle\mathrm{W}}, \underrightarrow{\mathcal{F}}_{\scriptscriptstyle\mathrm{B}}$: Global world and body frames in our estimator, where $\underrightarrow{\mathcal{F}}_{\scriptscriptstyle\mathrm{W}}$ is gravity-aligned and $\underrightarrow{\mathcal{F}}_{\scriptscriptstyle\mathrm{B}}$ also defines the global clock.
\item $\boldsymbol{T}^{\scriptscriptstyle\mathrm{B}}_{\scriptscriptstyle\mathrm{A}} \in SE\left(3\right)$: The homogeneous transformation from frame $\underrightarrow{\mathcal{F}}_{\scriptscriptstyle\mathrm{A}}$ to frame $\underrightarrow{\mathcal{F}}_{\scriptscriptstyle\mathrm{B}}$, which is composed of a rotation matrix $\boldsymbol{R}^{\scriptscriptstyle\mathrm{B}}_{\scriptscriptstyle\mathrm{A}} \in SO\left(3\right)$ and a translation vector $\boldsymbol{p}^{\scriptscriptstyle\mathrm{B}}_{\scriptscriptstyle\mathrm{A}} \in \mathbb{R}^3$.
\item $\boldsymbol{\omega}^{\scriptscriptstyle\mathrm{B}}_{\scriptscriptstyle\mathrm{A}}, \boldsymbol{a}^{\scriptscriptstyle\mathrm{B}}_{\scriptscriptstyle\mathrm{A}} \in \mathbb{R}^3$: The angular velocity and acceleration of frame $\underrightarrow{\mathcal{F}}_{\scriptscriptstyle\mathrm{A}}$ with respect to and parameterized in frame $\underrightarrow{\mathcal{F}}_{\scriptscriptstyle\mathrm{B}}$.
\end{itemize}

\subsection{Problem Statement}

The proposed estimator aims to tightly couple the pose measurements $\tilde{\boldsymbol{T}}_{\scriptscriptstyle\mathrm{M}}^{\scriptscriptstyle\mathrm{P}}$ from the MoCap system, the inertial data (i.e., angular velocity  $\tilde{\boldsymbol{\omega}}$ and acceleration $\tilde{\boldsymbol{a}}$) from the auxiliary IMU, and the poses $\tilde{\boldsymbol{T}}^{\scriptscriptstyle\mathrm{H}}_{\scriptscriptstyle\mathrm{D}}$ provided by the DUT. It performs spatiotemporal calibration of each sensor relative to the global coordinate frames $\underrightarrow{\mathcal{F}}_{\scriptscriptstyle\mathrm{W}}$ and $\underrightarrow{\mathcal{F}}_{\scriptscriptstyle\mathrm{B}}$, and simultaneously estimates the optimal trajectory $\boldsymbol{T}^{\scriptscriptstyle\mathrm{W}}_{\scriptscriptstyle\mathrm{B}} \left( t \right)$. In this paper, the global coordinate frame is treated as an independent reference frame. This design ensures compatibility with scenarios where the DUT provides IMU data, enabling multi-IMU fusion. In practice, the global frame can be configured to coincide with the frame of any input sensor. After optimization, the estimated trajectory is transformed to the DUT’s reference frame and clock using spatiotemporal extrinsic parameters, allowing it to serve as GT for SLAM benchmarking.

\begin{table}[]
\caption{Estimated states and their definitions in the proposed method. }
\label{TAB: Estimated states and their definitions}
\centering
\renewcommand{\arraystretch}{1.02}
\footnotesize
\begin{tabular}{c|p{6.2cm}}
\toprule[1.0pt]
 & Time-Varying \\ \midrule
$\boldsymbol{R}^{\scriptscriptstyle\mathrm{W}}_{\scriptscriptstyle\mathrm{B}} \left( t \right)$ &  Rotation of the global rigid-body \\
$\boldsymbol{p}^{\scriptscriptstyle\mathrm{W}}_{\scriptscriptstyle\mathrm{B}} \left( t \right)$ &   Translation of the global rigid-body \\ 
\midrule
$\Delta t_{\scriptscriptstyle\mathrm{M}}^{\scriptscriptstyle\mathrm{B}} \left( t \right)$ & Time offset from the MoCap clock to the global clock \\
$\Delta t_{\scriptscriptstyle\mathrm{I}}^{\scriptscriptstyle\mathrm{B}} \left( t \right)$ & Time offset from the IMU clock to the global clock \\
$\Delta t_{\scriptscriptstyle\mathrm{D}}^{\scriptscriptstyle\mathrm{B}} \left( t \right)$ & Time offset from the DUT clock to the global clock \\
\midrule
$\boldsymbol{b}_{\scriptscriptstyle\boldsymbol{\omega}} \left( t \right)$ & Gyroscope bias \\
$\boldsymbol{b}_{\scriptscriptstyle\boldsymbol{a}} \left( t \right)$ & Accelerometer bias \\
\midrule \midrule
 & Time-Invariant \\ 
 \midrule
$\boldsymbol{T}^{\scriptscriptstyle\mathrm{B}}_{\scriptscriptstyle\mathrm{M}}$ & Body frame transformation from MoCap to global \\
$\boldsymbol{T}^{\scriptscriptstyle\mathrm{B}}_{\scriptscriptstyle\mathrm{I}}$ & Body frame transformation from IMU to global \\
$\boldsymbol{T}^{\scriptscriptstyle\mathrm{B}}_{\scriptscriptstyle\mathrm{D}}$ & Body frame transformation from DUT to global \\
$\boldsymbol{T}^{\scriptscriptstyle\mathrm{W}}_{\scriptscriptstyle\mathrm{P}}$ & World frame transformation from MoCap to global \\  
\midrule
$\boldsymbol{R}^{\scriptscriptstyle\boldsymbol{\omega}}_{\scriptscriptstyle\boldsymbol{a}}$ & Misalignment between accelerometer and gyroscope \\
$\boldsymbol{M}_{\scriptscriptstyle\boldsymbol{\omega}}$ & Gyroscope scale and non-orthogonality \\
$\boldsymbol{M}_{\scriptscriptstyle\boldsymbol{a}}$ & Accelerometer scale and non-orthogonality \\ 
\bottomrule[1.0pt]
\end{tabular}
\vspace{-8pt}
\end{table}

The state variables estimated by our method are summarized in \cref{TAB: Estimated states and their definitions}, with categorization into time-varying and time-invariant states. The time-varying states are represented using continuous-time functions, which will be discussed in detail later. The time offset is used to transform the timestamp $\tau_{\scriptscriptstyle\mathrm{A}}$ from the clock $\underrightarrow{\mathcal{F}}_{\scriptscriptstyle\mathrm{A}}$ to the global clock $\underrightarrow{\mathcal{F}}_{\scriptscriptstyle\mathrm{B}}$, which is expressed as: $t_{\scriptscriptstyle\mathrm{A}}=\Delta t_{\scriptscriptstyle\mathrm{A}}^{\scriptscriptstyle\mathrm{B}}\left( t \right) + \tau_{\scriptscriptstyle\mathrm{A}}$. To fully leverage the IMU data, we also estimate its intrinsic parameters. Specifically, the IMU bias is modeled as a random walk \cite{furgale2013unified}.  The rotational misalignment between the accelerometer and gyroscope is accounted for using the matrix $\boldsymbol{R}^{\scriptscriptstyle\boldsymbol{\omega}}_{\scriptscriptstyle\boldsymbol{a}}$, assuming that the IMU body frame $\underrightarrow{\mathcal{F}}_{\scriptscriptstyle\mathrm{I}}$ aligns with the accelerometer frame $\underrightarrow{\mathcal{F}}_{\scriptscriptstyle\mathrm{a}}$. Additionally, $\boldsymbol{M}_{\scriptscriptstyle\boldsymbol{\omega}}$ and $\boldsymbol{M}_{\scriptscriptstyle\boldsymbol{a}}$ are upper triangular matrices, whose diagonal elements represent scale factors, and the off-diagonal elements account for non-orthogonality effects \cite{chen2024ikalibr}.

\section{Methodology} \label{SEC: Methodology}

\begin{figure}[t]
\centering
\includegraphics[width=0.9\linewidth]{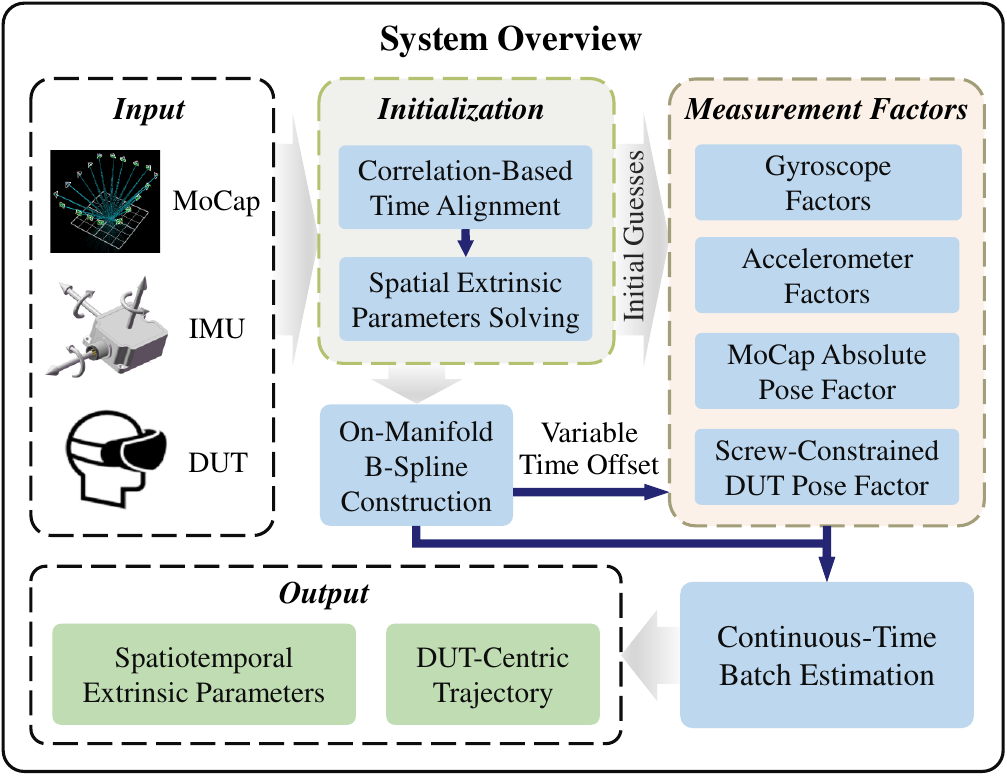}
\caption{Overview of the proposed estimator, which adopts a coarse-to-fine strategy and uses continuous-time batch estimation to perform spatiotemporal calibration and achieve localization trajectory estimation.}
\label{FIG: System overview}
\vspace{-10pt}
\end{figure}

The framework of the proposed system is illustrated in \cref{FIG: System overview}.  It starts with initializing spatiotemporal calibration parameters as initial guesses for nonlinear optimization. Then, B-spline functions parameterize time-varying states, including kinematics and variable time offsets. Measurement factors are then formulated by integrating sensor measurements with their corresponding models. The complementary strengths of MoCap and IMU are leveraged to reduce errors, and the screw constraint is employed to mitigate the impact of the accumulated errors from DUT measurements. Finally, continuous-time batch estimation refines calibration and outputs a high-precision DUT-centric trajectory result. 

\subsection{Initialization} \label{SEC: Initialization}

To ensure global convergence of the nonlinear optimization, a linear initialization of calibration parameters is required. The first critical step is temporal alignment, where we initially assume a constant time offset and employ a method based on cross-correlation \cite{pachtrachai2018chess} for estimation. Specifically, rigidly connected body frames share the same rotational motion, which means they exhibit identical angular velocities. This enables time synchronization through cross-correlation analysis, where a higher cross-correlation value indicates a stronger similarity between the signals. The IMU directly provides angular velocity measurements, whereas for pose measurements, angular velocity can be obtained by computing the first-order derivative of the rotation. Therefore, synchronization across all sensors is uniformly formulated as:
\begin{equation}
t_{\mathrm{shift}} = \underset{\mathrm{index}}{\operatorname{argmax}} \left(\operatorname{Corr}\left(\boldsymbol{\tilde{\omega}}_{\scriptscriptstyle{1}}, \boldsymbol{\tilde{\omega}}_{\scriptscriptstyle{2}}\right)\right),
\end{equation}
where $\operatorname{Corr}\left( \cdot \right)$ denotes the cross-correlation function, and $\boldsymbol{\tilde{\omega}}_{\scriptscriptstyle{1}}$ and $\boldsymbol{\tilde{\omega}}_{\scriptscriptstyle{2}}$ denote the angular velocity signals of any two sensors involved in the process. The time offset $\Delta t_{\scriptscriptstyle{2}}^{\scriptscriptstyle{1}}$ can be derived by converting the maximum cross-correlation index $t_{\mathrm{shift}}$ along with the velocity sampling interval.

Subsequently, the initial spatial calibration of all input sensors is performed. Depending on the sensor data type, spatial calibration can be categorized into pose-to-pose and pose-to-IMU calibration. For the former, the problem can be formulated as the classic $\boldsymbol{AX} = \boldsymbol{XB}$ hand-eye calibration, and solved using a linear least-squares approach as described in \cite{furrer2018evaluation}. For the latter, inspired by the hand-eye calibration, a two-step solution can be achieved based on the preintegration information \cite{qin2018vins} from the IMU.

Taking the MoCap and IMU calibration as an example, the rotation $\boldsymbol{R}^{\scriptscriptstyle\mathrm{M}}_{\scriptscriptstyle\mathrm{I}}$ between the body frames satisfies: 
\begin{equation}
\left(\left( \boldsymbol{\tilde{R}}^{\scriptscriptstyle\mathrm{P}}_{\scriptscriptstyle\mathrm{M}_{\scriptscriptstyle{i}}} \right)^{-1}  \boldsymbol{\tilde{R}}^{\scriptscriptstyle\mathrm{P}}_{\scriptscriptstyle\mathrm{M}_{\scriptscriptstyle{j}}} \right) \boldsymbol{R}^{\scriptscriptstyle\mathrm{M}}_{\scriptscriptstyle\mathrm{I}} = \boldsymbol{R}^{\scriptscriptstyle\mathrm{M}}_{\scriptscriptstyle\mathrm{I}} \boldsymbol{\tilde{R}}^{\scriptscriptstyle\mathrm{I}_{\scriptscriptstyle{i}}}_{\scriptscriptstyle\mathrm{I}_{\scriptscriptstyle{j}}},
\end{equation}
where $\boldsymbol{\tilde{R}}^{\scriptscriptstyle\mathrm{P}}_{\scriptscriptstyle\mathrm{M}_{\scriptscriptstyle{i}}}$ and $\boldsymbol{\tilde{R}}^{\scriptscriptstyle\mathrm{P}}_{\scriptscriptstyle\mathrm{M}_{\scriptscriptstyle{j}}}$ are MoCap measurements at two time instants $\delta t$ apart, and $\boldsymbol{\tilde{R}}^{\scriptscriptstyle\mathrm{I}_{\scriptscriptstyle{i}}}_{\scriptscriptstyle\mathrm{I}_{\scriptscriptstyle{j}}}$ denotes the corresponding rotational preintegration over $\delta t$. Converting the rotation into quaternion form and applying the left and right multiplication operators $\left[ \cdot \right]_{\scriptscriptstyle\mathrm{L}}$ and $\left[ \cdot \right]_{\scriptscriptstyle\mathrm{R}}$ yields a simple linear expression:
\begin{equation}
\left[ \left[ \boldsymbol{\tilde{q}}_{\scriptscriptstyle\mathrm{M}_j}^{\scriptscriptstyle\mathrm{M}_i} \right]_{\scriptscriptstyle\mathrm{L}} - \left[ \boldsymbol{\tilde{q}}_{\scriptscriptstyle\mathrm{I}_j}^{\scriptscriptstyle\mathrm{I}_{i}} \right]_{\scriptscriptstyle\mathrm{R}} \right]
 \boldsymbol{q}^{\scriptscriptstyle\mathrm{M}}_{\scriptscriptstyle\mathrm{I}} = \boldsymbol{0}.
\label{EQ: Rotational extrinsic initialization}
\end{equation}

Furthermore, the translational extrinsic parameters $\boldsymbol{p}^{\scriptscriptstyle\mathrm{M}}_{\scriptscriptstyle\mathrm{I}}$ between the body frames need to be obtained, along with gravity alignment vector $\boldsymbol{g}^{\scriptscriptstyle\mathrm{P}}$ for the world frame. Under rigid body motion constraints, the following equation holds: 
\begin{equation}
\boldsymbol{\tilde{R}}^{\scriptscriptstyle\mathrm{P}}_{\scriptscriptstyle\mathrm{I}_i} = \boldsymbol{\tilde{R}}^{\scriptscriptstyle\mathrm{P}}_{\scriptscriptstyle\mathrm{M}_i} \boldsymbol{R}^{\scriptscriptstyle\mathrm{M}}_{\scriptscriptstyle\mathrm{I}}, \ \
\boldsymbol{\tilde{p}}^{\scriptscriptstyle\mathrm{P}}_{\scriptscriptstyle\mathrm{I}_i} = \boldsymbol{\tilde{R}}^{\scriptscriptstyle\mathrm{P}}_{\scriptscriptstyle\mathrm{M}_i} \boldsymbol{p}^{\scriptscriptstyle\mathrm{M}}_{\scriptscriptstyle\mathrm{I}} + \boldsymbol{\tilde{p}}^{\scriptscriptstyle\mathrm{P}}_{\scriptscriptstyle\mathrm{M}_i}.
\end{equation}
Integrating these constraints with translational and velocity preintegration terms $\boldsymbol{\tilde{\alpha}}^{\scriptscriptstyle\mathrm{I}_i}_{\scriptscriptstyle\mathrm{I}_j}$ and $\boldsymbol{\tilde{\beta}}^{\scriptscriptstyle\mathrm{I}_i}_{\scriptscriptstyle\mathrm{I}_j}$ yields the following linear equation:
\begin{equation}
\boldsymbol{A}^{\scriptscriptstyle{i}}_{\scriptscriptstyle{j}}
\left[
\begin{matrix}
\boldsymbol{p}^{\scriptscriptstyle\mathrm{M}}_{\scriptscriptstyle\mathrm{I}} \\
\boldsymbol{g}^{\scriptscriptstyle\mathrm{P}} \\
\boldsymbol{v}^{\scriptscriptstyle\mathrm{P}}_{\scriptscriptstyle\mathrm{I}_i} \\
\boldsymbol{v}^{\scriptscriptstyle\mathrm{P}}_{\scriptscriptstyle\mathrm{I}_j}
\end{matrix}
\right] = 
\left[ 
\begin{matrix}
\boldsymbol{\tilde{R}}^{\scriptscriptstyle\mathrm{P}}_{\scriptscriptstyle\mathrm{M}_i} \boldsymbol{R}^{\scriptscriptstyle\mathrm{M}}_{\scriptscriptstyle\mathrm{I}} \boldsymbol{\tilde{\alpha}}^{\scriptscriptstyle\mathrm{I}_i}_{\scriptscriptstyle\mathrm{I}_j} + \boldsymbol{\tilde{p}}^{\scriptscriptstyle\mathrm{P}}_{\scriptscriptstyle\mathrm{M}_i} - \boldsymbol{\tilde{p}}^{\scriptscriptstyle\mathrm{P}}_{\scriptscriptstyle\mathrm{M}_j}\\
\boldsymbol{\tilde{R}}^{\scriptscriptstyle\mathrm{P}}_{\scriptscriptstyle\mathrm{M}_i} \boldsymbol{R}^{\scriptscriptstyle\mathrm{M}}_{\scriptscriptstyle\mathrm{I}} \boldsymbol{\tilde{\beta}}^{\scriptscriptstyle\mathrm{I}_i}_{\scriptscriptstyle\mathrm{I}_j}
\end{matrix} 
\right], 
\label{EQ: Translational extrinsic initialization}
\end{equation}
where $\boldsymbol{v}^{\scriptscriptstyle\mathrm{P}}_{\scriptscriptstyle\mathrm{I}}$ denotes the additional intermediate velocity terms, and the coefficient matrix is given by
\begin{equation}
\boldsymbol{A}^{\scriptscriptstyle{i}}_{\scriptscriptstyle{j}} = 
\left[ 
\begin{matrix}
-\boldsymbol{\tilde{R}}^{\scriptscriptstyle\mathrm{P}}_{\scriptscriptstyle\mathrm{M}_i} + \boldsymbol{\tilde{R}}^{\scriptscriptstyle\mathrm{P}}_{\scriptscriptstyle\mathrm{M}_j} & \frac{1}{2} \delta t^2 & -\boldsymbol{I}\delta t & \boldsymbol{0} \\
\boldsymbol{0} & \delta t & -\boldsymbol{I} & \boldsymbol{I} 
\end{matrix}
\right]. 
\end{equation}

For a given motion sequence, the constraints in \cref{EQ: Rotational extrinsic initialization} and \cref{EQ: Translational extrinsic initialization} can be stacked to obtain linear least-squares solutions, enabling the initialization of the pose-to-IMU spatial extrinsic parameters. Additionally, since the intrinsic parameters of the IMU have small magnitudes in our study, they can be directly initialized as identity matrices or zeros. At this stage, the initialization of the time offset and all time-invariant states summarized in \cref{TAB: Estimated states and their definitions} has been completed. 

\subsection{Continuous-Time Representation}

In this work, a continuous-time parameterization is employed for time-varying state variables. This representation enables closed-form sampling at arbitrary timestamps, ensuring temporal differentiability, making it an ideal choice for the efficient fusion of high-frequency, asynchronous multi-sensor measurements. Among various continuous-time implementations, we follow the B-spline formulation in \cite{sommer2020efficient} and separately construct $\left( i \right)$ B-splines in Euclidean space to encode $n$-dimensional vectors, including translational kinematics, IMU biases, and temporal offsets, and $\left( ii \right)$ B-splines on the $SO\left(3\right)$ manifold to encode rotational kinematics.

Specifically, given $n$-dimensional Euclidean space control points $\boldsymbol{p}$ uniformly distributed in the time domain along with their corresponding timestamps $\tau$:
\begin{equation}
\mathcal{P} = \left \{ \boldsymbol{p}_{\scriptscriptstyle{i}}, \tau_{\scriptscriptstyle{i}} | \boldsymbol{p}_{\scriptscriptstyle{i}} \in \mathbb{R}^{\scriptscriptstyle{n}}, \tau_{\scriptscriptstyle{i}} \in \mathbb{R}, i \in \mathbb{N} \right \} \quad s.t.\quad  \tau_{\scriptscriptstyle{i+1}}-\tau_{\scriptscriptstyle{i}} \equiv \Delta \tau_{\scriptscriptstyle\boldsymbol{p}},
\label{EQ: B-spline control points}
\end{equation}
a $k$-order B-spline in $\mathbb{R}^{\scriptscriptstyle{n}}$ at time $t \in \left[ \tau_{\scriptscriptstyle{i}}, \tau_{\scriptscriptstyle{i+1}} \right)$ will involve $k$ control points and be expressed in cumulative form as
\begin{equation}
\boldsymbol{p} \left ( t \right ) = \boldsymbol{p}_{\scriptscriptstyle{i}} + \sum_{j=1}^{k-1}B_{\scriptscriptstyle{j}}\left ( u \right )\left ( \boldsymbol{p}_{\scriptscriptstyle{i+j}} - \boldsymbol{p}_{\scriptscriptstyle{i+j-1}} \right ), 
\end{equation}
where $u = \left ( t - \tau_{\scriptscriptstyle{i}} \right ) / \Delta \tau_{\scriptscriptstyle\boldsymbol{p}}$, and $B_{\scriptscriptstyle{j}}\left ( u \right )$ denotes the $j$-th element of the vector $\boldsymbol{B}\left ( u \right )$, which is computed as the matrix product of the order-dependent cumulative matrix $\boldsymbol{M}_{\left ( k \times k \right )}$ and the normalized time vector $\boldsymbol{U}$ as 
\begin{equation}
\boldsymbol{B}\left ( u \right ) = \boldsymbol{M}_{\left ( k \times k \right )} \boldsymbol{U}, \ \ \mathrm{where} \ \ \boldsymbol{U} = \begin{bmatrix}
u^{\scriptscriptstyle{0}} & u^{\scriptscriptstyle{1}} & \dots & u^{\scriptscriptstyle{k-1}}
\end{bmatrix}^{\scriptscriptstyle\top}.
\label{EQ: Cumulative matrix and normalized tiem vector}
\end{equation}

The control points of a rotational B-spline are distributed on the $SO\left(3\right)$ manifold, following a form similar to \cref{EQ: B-spline control points}. Due to the non-closedness of scalar multiplication on the $SO\left(3\right)$ manifold, accumulating control points requires forward and inverse mapping of the tangent space. A $k$-order rotational B-spline is given by: 
\begin{equation}
\boldsymbol{R} \left ( t \right ) = \boldsymbol{R}_{\scriptscriptstyle{i}} \prod_{j=1}^{k-1}\mathrm{Exp} \left ( B_{\scriptscriptstyle{j}}\left ( u \right )\mathrm{Log}\left ( \boldsymbol{R}_{\scriptscriptstyle{i+j-1}}^{\scriptscriptstyle\top} \boldsymbol{R}_{\scriptscriptstyle{i+j}} \right ) \right ),
\end{equation}
where $\mathrm{Exp}\left( \cdot \right)$ denotes the exponential map from the Lie algebra to the Lie group,  $\mathrm{Log}\left( \cdot \right)$ represents its inverse process.

\begin{figure}[t]
\centering
\includegraphics[width=0.95\linewidth]{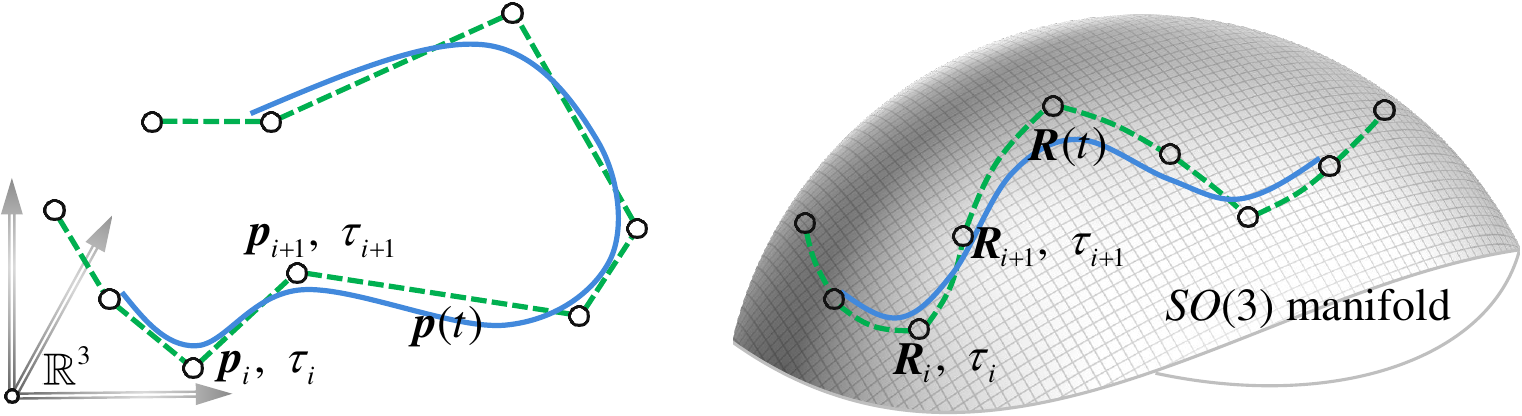}
\caption{Illustration of B-splines in $\mathbb{R}^{\scriptscriptstyle{3}}$ and on the $SO\left(3\right)$ manifold. Black circles denote control points, and the blue curve represents the cubic B-spline. The linear B-spline is shown as a green dashed line, appearing as a straight line in $\mathbb{R}^{\scriptscriptstyle{3}}$ and a geodesic on $SO(3)$.}
\label{FIG: Spline visualization}
\vspace{-10pt}
\end{figure}

In this work, to achieve $\mathcal{C}^{\scriptscriptstyle{2}}$-continuous motion modeling, cubic B-splines ($k=4$) are used to encode the kinematic state, as shown in \cref{FIG: Spline visualization}. This approach ensures smooth transitions in translation, velocity, and acceleration. The time offset $\Delta t$ is also treated as a time-varying state variable in our work to account for the time drift between independent clocks. We propose the use of B-splines for unified modeling. To avoid unnecessary continuity constraints from higher-order B-splines, we choose a linear approach ($k=2$). This representation is similar to the green dashed line on the left side of \cref{FIG: Spline visualization}, where the control points lie in $\mathbb{R}$ space and the spline curve hits them directly. Subsequently, we can hierarchically implement the function for temporal calibration. As an example, for a MoCap measurement with timestamp $\tau_{\scriptscriptstyle{\mathbf{M}_{i}}}$, the following formulation samples the global rigid-body motion to construct residuals:
\begin{equation}
\boldsymbol{R}^{\scriptscriptstyle\mathrm{W}}_{\scriptscriptstyle\mathrm{B}_i} = \boldsymbol{R}^{\scriptscriptstyle\mathrm{W}}_{\scriptscriptstyle\mathrm{B}} \left( t_{\scriptscriptstyle\mathrm{M}_{i}} \right), \ \
\boldsymbol{p}^{\scriptscriptstyle\mathrm{W}}_{\scriptscriptstyle\mathrm{B}_i} = \boldsymbol{p}^{\scriptscriptstyle\mathrm{W}}_{\scriptscriptstyle\mathrm{B}} \left( t_{\scriptscriptstyle\mathrm{M}_{i}} \right), \ \ \mathrm{where}  \ \ t_{\scriptscriptstyle\mathrm{M}_{i}} = \Delta t_{\scriptscriptstyle\mathrm{M}}^{\scriptscriptstyle\mathrm{B}}\left( \tau_{\scriptscriptstyle{\mathbf{M}_{i}}} \right) + \tau_{\scriptscriptstyle{\mathbf{M}_{i}}}.
\end{equation}
This method will be used in all subsequent derivations, enabling gradient computation for the control points distributed in the time domain to estimate variable time offsets. Notably, the interval of the time offset control points needs to be carefully considered in practice, primarily depending on clock accuracy and the adequacy of motion excitation. In our implementation, we adopted a 20-second interval, which ensures optimization convergence and precise estimation of clock drift.

For the rigid-body motion B-spline functions $\boldsymbol{R}^{\scriptscriptstyle\mathrm{W}}_{\scriptscriptstyle\mathrm{B}} \left( t \right)$ and $\boldsymbol{p}^{\scriptscriptstyle\mathrm{W}}_{\scriptscriptstyle\mathrm{B}} \left( t \right)$, control points can be directly initialized using MoCap measurements, and then transformed according to the initialized spatiotemporal extrinsic parameters. For other B-spline functions, we initialize them to constant values based on the results in \cref{SEC: Initialization}. Subsequently, these continuous-time representations serve as the basis for formulating measurement factors and performing batch estimation.

\subsection{Pose Factors} \label{SEC: Pose Factor}

This section first derives the pose factors for the MoCap system and the DUT. The MoCap system provides pose measurements $\tilde{\boldsymbol{R}}^{\scriptscriptstyle\mathrm{P}}_{\scriptscriptstyle\mathrm{M}}$ and $\tilde{\boldsymbol{p}}^{\scriptscriptstyle\mathrm{P}}_{\scriptscriptstyle\mathrm{M}}$, and the predicted values, referenced to the MoCap frame, can be derived from the system states at the given timestamp $t_{\scriptscriptstyle\mathrm{M}_{i}}$:
\begin{align}
\boldsymbol{R}^{\scriptscriptstyle\mathrm{P}}_{\scriptscriptstyle\mathrm{M}_{i}} &=  \boldsymbol{R}^{\scriptscriptstyle\mathrm{P}}_{\scriptscriptstyle\mathrm{W}} \boldsymbol{R}^{\scriptscriptstyle\mathrm{W}}_{\scriptscriptstyle\mathrm{B}}  \left( t_{\scriptscriptstyle\mathrm{M}_{i}} \right) \boldsymbol{R}^{\scriptscriptstyle\mathrm{B}}_{\scriptscriptstyle\mathrm{M}}, \\
\boldsymbol{p}^{\scriptscriptstyle\mathrm{P}}_{\scriptscriptstyle\mathrm{M}_{i}} &= \boldsymbol{R}^{\scriptscriptstyle\mathrm{P}}_{\scriptscriptstyle\mathrm{W}} \boldsymbol{R}^{\scriptscriptstyle\mathrm{W}}_{\scriptscriptstyle\mathrm{B}} \left( t_{\scriptscriptstyle\mathrm{M}_{i}} \right)\boldsymbol{p}^{\scriptscriptstyle\mathrm{B}}_{\scriptscriptstyle\mathrm{M}} + \boldsymbol{R}^{\scriptscriptstyle\mathrm{P}}_{\scriptscriptstyle\mathrm{W}} \boldsymbol{p}^{\scriptscriptstyle\mathrm{W}}_{\scriptscriptstyle\mathrm{B}} \left( t_{\scriptscriptstyle\mathrm{M}_{i}} \right) + \boldsymbol{p}^{\scriptscriptstyle\mathrm{P}}_{\scriptscriptstyle\mathrm{W}}. 
\end{align}
The corresponding MoCap residual is given by:
\begin{equation}
\boldsymbol{r}_{\scriptscriptstyle\mathrm{M}} \left ( \tilde{\boldsymbol{T}}^{\scriptscriptstyle\mathrm{P}}_{\scriptscriptstyle\mathrm{M}_{i}}, \mathcal{X} \right ) = \begin{bmatrix}
\boldsymbol{r}_{\scriptscriptstyle{\boldsymbol{R}}} \\
\boldsymbol{r}_{\scriptscriptstyle{\boldsymbol{p}}}
\end{bmatrix} = 
\begin{bmatrix}
\mathrm{Log} \left ( \left ( \tilde{\boldsymbol{R}}^{\scriptscriptstyle\mathrm{P}}_{\scriptscriptstyle\mathrm{M}_{i}} \right )^{\scriptscriptstyle{-1}} \boldsymbol{R}^{\scriptscriptstyle\mathrm{P}}_{\scriptscriptstyle\mathrm{M}_{i}} \right )   \\
\boldsymbol{p}^{\scriptscriptstyle\mathrm{P}}_{\scriptscriptstyle\mathrm{M}_{i}} - \tilde{\boldsymbol{p}}^{\scriptscriptstyle\mathrm{P}}_{\scriptscriptstyle\mathrm{M}_{i}}
\end{bmatrix},
\end{equation}
where $\mathcal{X}$ uniformly represents all involved system state variables. The rotation error is mapped to the tangent space to quantify the error. Despite the presence of some jitter, the MoCap system provides absolute pose constraints without cumulative errors, which forms one of the key foundations for enabling the estimator to achieve high-precision trajectory estimation.

Unlike MoCap systems, the measurements from the DUT exhibit unpredictable drift. To mitigate the impact of accumulated errors, an optimal strategy is to represent the DUT poses over a time interval in a relative form. Specifically, for measurements at timestamps $t_{\scriptscriptstyle\mathrm{D}} \in [ t_{\scriptscriptstyle\mathrm{D}_{i}}, t_{\scriptscriptstyle\mathrm{D}_{j}} ]$, the predicted value can be expressed as
\begin{equation}
\begin{bmatrix}
\boldsymbol{R}^{\scriptscriptstyle\mathrm{D}_{i}}_{\scriptscriptstyle\mathrm{D}_{j}} & \boldsymbol{p}^{\scriptscriptstyle\mathrm{D}_{i}}_{\scriptscriptstyle\mathrm{D}_{j}} \\
\boldsymbol{0} & 1
\end{bmatrix} = 
\boldsymbol{T}^{\scriptscriptstyle\mathrm{D}_{i}}_{\scriptscriptstyle\mathrm{D}_{j}} = 
\left ( \boldsymbol{T}^{\scriptscriptstyle\mathrm{B}}_{\scriptscriptstyle\mathrm{D}} \right ) ^ {\scriptscriptstyle{-1} }
\left ( \boldsymbol{T}^{\scriptscriptstyle\mathrm{W}}_{\scriptscriptstyle\mathrm{B}} \left( t_{\scriptscriptstyle\mathrm{M}_{i}} \right) \right ) ^ {\scriptscriptstyle{-1} }
\boldsymbol{T}^{\scriptscriptstyle\mathrm{W}}_{\scriptscriptstyle\mathrm{B}} \big( t_{\scriptscriptstyle\mathrm{M}_{j}} \big)
\boldsymbol{T}^{\scriptscriptstyle\mathrm{B}}_{\scriptscriptstyle\mathrm{D}}.
\end{equation}
For conciseness, the homogeneous transformation representation is used, where the rotation and translation components can be extracted from the corresponding submatrices. The corresponding relative pose residual for the DUT is formulated as follows:
\begin{equation}
\boldsymbol{r}_{\scriptscriptstyle\mathrm{D}} 
\left ( \tilde{\boldsymbol{T}}^{\scriptscriptstyle\mathrm{H}}_{\scriptscriptstyle\mathrm{D}_{i}}, 
\tilde{\boldsymbol{T}}^{\scriptscriptstyle\mathrm{H}}_{\scriptscriptstyle\mathrm{D}_{j}}, 
\mathcal{X} \right ) = 
\begin{bmatrix}
\boldsymbol{r}_{\scriptscriptstyle{\boldsymbol{R}}} \\
\boldsymbol{r}_{\scriptscriptstyle{\boldsymbol{p}}}
\end{bmatrix} = 
\begin{bmatrix}
\mathrm{Log} \left (
\left ( \tilde{\boldsymbol{R}}^{\scriptscriptstyle\mathrm{H}}_{\scriptscriptstyle\mathrm{D}_{j}} \right ) ^ {\scriptscriptstyle{{-1}} } 
\tilde{\boldsymbol{R}}^{\scriptscriptstyle\mathrm{H}}_{\scriptscriptstyle\mathrm{D}_{i}} 
\boldsymbol{R}^{\scriptscriptstyle\mathrm{D}_{i}}_{\scriptscriptstyle\mathrm{D}_{j}}
\right ) \\
\boldsymbol{p}^{\scriptscriptstyle\mathrm{D}_{i}}_{\scriptscriptstyle\mathrm{D}_{j}} - 
\left ( \tilde{\boldsymbol{R}}^{\scriptscriptstyle\mathrm{H}}_{\scriptscriptstyle\mathrm{D}_{i}} \right ) ^ {\scriptscriptstyle{{-1}} } 
\left ( \tilde{\boldsymbol{p}}^{\scriptscriptstyle\mathrm{H}}_{\scriptscriptstyle\mathrm{D}_{j}} - 
\tilde{\boldsymbol{p}}^{\scriptscriptstyle\mathrm{H}}_{\scriptscriptstyle\mathrm{D}_{i}}\right )
\end{bmatrix},
\end{equation}
This residual term is designed to have zero gradient with respect to $\boldsymbol{T}^{\scriptscriptstyle\mathrm{W}}_{\scriptscriptstyle\mathrm{B}}  \left( t \right)$, ensuring it is used solely for spatiotemporal calibration without affecting the global motion trajectory. Moreover, the use of relative poses eliminates the need to estimate the transformation $\boldsymbol{T}^{\scriptscriptstyle\mathrm{W}}_{\scriptscriptstyle\mathrm{H}}$ of the world frame.

Notably, to ensure calibration accuracy, the selection of the interval for relative pose measurements and the weighting of residual terms are critical considerations.
Based on screw theory, when expressing the same rigid-body motion in different coordinate frames, the following constraint is satisfied: the rotation angle $\theta$ and the translation $d$ along the screw axis remain equal, known as the screw congruence theorem \cite{pachtrachai2018chess}. Since the global motion trajectory $\boldsymbol{T}^{\scriptscriptstyle\mathrm{W}}_{\scriptscriptstyle\mathrm{B}}$ is constrained by MoCap with high accuracy, we use it as a reference to design the weighting function $w$ for the DUT residuals. Specifically, the rotational component $w_{\scriptscriptstyle{\boldsymbol{R}}}$ is given by
\begin{equation}
w_{\scriptscriptstyle{\boldsymbol{R}}}\left ( 
\theta^{\scriptscriptstyle\mathrm{D}_{i}}_{\scriptscriptstyle\mathrm{D}_{j}}, 
\theta^{\scriptscriptstyle\mathrm{B}_{i}}_{\scriptscriptstyle\mathrm{B}_{j}} \right ) = 
\mathrm{exp} \left ( -\frac{\theta^{\scriptscriptstyle\mathrm{D}_{i}}_{\scriptscriptstyle\mathrm{D}_{j}} - 
\theta^{\scriptscriptstyle\mathrm{B}_{i}}_{\scriptscriptstyle\mathrm{B}_{j}}}
{\theta^{\scriptscriptstyle\mathrm{B}_{i}}_{\scriptscriptstyle\mathrm{B}_{j}}}  \right ) ^ {\scriptscriptstyle{2}}.
\end{equation}
A Gaussian-inspired weighting function attenuates residuals deviating from the screw constraint. Similarly, for the translational component $w_{\scriptscriptstyle{\boldsymbol{p}}}$, $d$ is used as a reference, using the same formulation. The values of $\theta$ and $d$ can be derived based on the inverse Rodrigues formula \cite{dai2015euler}:
\begin{equation}
\theta = \mathrm{arccos} \left ( \frac{\mathrm{tr}\left ( \boldsymbol{R} \right ) -1}{2} \right ), \ 
d = \frac{ \left ( \boldsymbol{R} - \boldsymbol{R}^{\scriptscriptstyle\top} \right ) ^ {\scriptscriptstyle{\vee}}
\cdot \boldsymbol{p} } 
{2\mathrm{sin} \theta},
\end{equation}
where $\mathrm{tr}\left(\cdot\right)$ denotes the trace of a matrix, and $\left(\cdot\right)^{\scriptscriptstyle\vee}$ represents the vee map. In terms of relative pose intervals, it is essential to ensure sufficient motion excitation while preventing accumulated errors caused by long durations. Rotational excitation $\theta$, translational excitation $d$, and the time interval $\delta t_{\scriptscriptstyle{\mathrm{D}}}$ are simultaneously considered. In this work, we set the thresholds for these indicators as 5°, 0.1 m, and 0.5 s based on experimental considerations. A relative pose residual is constructed whenever any of these indicators exceeds its corresponding threshold.

\subsection{IMU Factors} \label{SEC: IMU Factors}

The IMU factor can be decomposed into two components: the gyroscope and the accelerometer. For the gyroscope factor, the ideal angular velocity measurement at time $t_{\scriptscriptstyle\mathrm{I}_{i}}$ is obtained by transforming the first derivative of the global rotation B-spline into the frame $\underrightarrow{\mathcal{F}}_{\scriptscriptstyle\mathrm{I}}$:
\begin{equation}
\boldsymbol{\omega}_{\scriptscriptstyle\mathrm{I}_{i}} = 
\left ( \boldsymbol{R} ^{\scriptscriptstyle\mathrm{W}}_{\scriptscriptstyle\mathrm{B}} \left( t_{\scriptscriptstyle\mathrm{I}_{i}} \right) 
\boldsymbol{R}^{\scriptscriptstyle\mathrm{B}}_{\scriptscriptstyle\mathrm{I}} 
\right ) ^ {\scriptscriptstyle{-1} }
\left ( \dot{\boldsymbol{R}} ^{\scriptscriptstyle\mathrm{W}}_{\scriptscriptstyle\mathrm{B}} \left( t_{\scriptscriptstyle\mathrm{I}_{i}} \right)
 \right ) ^ {\scriptscriptstyle{\vee} },
\end{equation}
Incorporating the gyroscope intrinsics from \cref{TAB: Estimated states and their definitions}, the residual is formulated as
\begin{equation}
\boldsymbol{r}_{\scriptscriptstyle\mathrm{I,\ \boldsymbol\omega}}
\left ( \tilde{\boldsymbol{\omega}}_{\scriptscriptstyle\mathrm{I}_{i}}, 
\mathcal{X} \right )  = 
\boldsymbol{M}_{\scriptscriptstyle\boldsymbol{\omega}} 
\boldsymbol{R}^{\scriptscriptstyle\boldsymbol{\omega}}_{\scriptscriptstyle\boldsymbol{a}}
\boldsymbol{\omega}_{\scriptscriptstyle\mathrm{I}_{i}} + 
\boldsymbol{b}_{\scriptscriptstyle\boldsymbol{\omega}} 
\left( t_{\scriptscriptstyle\mathrm{I}_{i}} \right) - 
\tilde{\boldsymbol{\omega}}_{\scriptscriptstyle\mathrm{I}_{i}}.
\end{equation}

The acceleration of the body frame $\underrightarrow{\mathcal{F}}_{\scriptscriptstyle\mathrm{I}}$ in the world frame $\underrightarrow{\mathcal{F}}_{\scriptscriptstyle\mathrm{W}}$ is transformed to obtain the ideal accelerometer measurement at time $t_{\scriptscriptstyle\mathrm{I}_{i}}$: 
\begin{equation}
\boldsymbol{a}_{\scriptscriptstyle\mathrm{I}_{i}} = 
\left ( \boldsymbol{R} ^{\scriptscriptstyle\mathrm{W}}_{\scriptscriptstyle\mathrm{B}} \left( t_{\scriptscriptstyle\mathrm{I}_{i}} \right) 
\boldsymbol{R}^{\scriptscriptstyle\mathrm{B}}_{\scriptscriptstyle\mathrm{I}} \right ) ^ {\scriptscriptstyle{-1} }
\left ( \boldsymbol{a}_{\scriptscriptstyle\mathrm{I}}^{\scriptscriptstyle\mathrm{W}} 
\left( t_{\scriptscriptstyle\mathrm{I}_{i}} \right) - 
\boldsymbol{g}^{\scriptscriptstyle\mathrm{W}} \right ),
\end{equation}
where $\boldsymbol{g}^{\scriptscriptstyle\mathrm{W}}$ is the gravity vector, defined in the gravity-aligned global frame $\underrightarrow{\mathcal{F}}_{\scriptscriptstyle\mathrm{W}}$, having only a $z$-component. The acceleration $\boldsymbol{a}_{\scriptscriptstyle\mathrm{I}}^{\scriptscriptstyle\mathrm{W}} \left( t_{\scriptscriptstyle\mathrm{I}_{i}} \right)$ can be obtained from the second derivative of the global motion B-spline:
\begin{equation}
\boldsymbol{a}_{\scriptscriptstyle\mathrm{I}} ^ {\scriptscriptstyle\mathrm{W}}
\left( t_{\scriptscriptstyle\mathrm{I}_{i}} \right) = 
\ddot{\boldsymbol{p}}_{\scriptscriptstyle\mathrm{I}}^{\scriptscriptstyle\mathrm{W}} 
\left( t_{\scriptscriptstyle\mathrm{I}_{i}} \right) = 
\ddot{\boldsymbol{R}}^{\scriptscriptstyle\mathrm{W}}_{\scriptscriptstyle\mathrm{B}}
\left( t_{\scriptscriptstyle\mathrm{I}_{i}} \right )
\boldsymbol{p}^{\scriptscriptstyle\mathrm{B}}_{\scriptscriptstyle\mathrm{I}} + 
\ddot{\boldsymbol{p}}^{\scriptscriptstyle\mathrm{W}}_{\scriptscriptstyle\mathrm{B}} 
\left( t_{\scriptscriptstyle\mathrm{I}_{i}} \right ).
\end{equation}
Accounting for the accelerometer intrinsic parameters, the residual is further formulated as
\begin{equation}
\boldsymbol{r}_{\scriptscriptstyle\mathrm{I,\ \boldsymbol{a}}}
\left ( \tilde{\boldsymbol{a}}_{\scriptscriptstyle\mathrm{I}_{i}}, 
\mathcal{X} \right ) = 
\boldsymbol{M}_{\scriptscriptstyle\boldsymbol{a}}
\boldsymbol{a}_{\scriptscriptstyle\mathrm{I}_{i}} + 
\boldsymbol{b}_{\scriptscriptstyle\boldsymbol{a}} \left( t \right) - 
\tilde{\boldsymbol{a}}_{\scriptscriptstyle\mathrm{I}_{i}}.
\end{equation}

\begin{figure}[t]
\centering
\includegraphics[width=\linewidth]{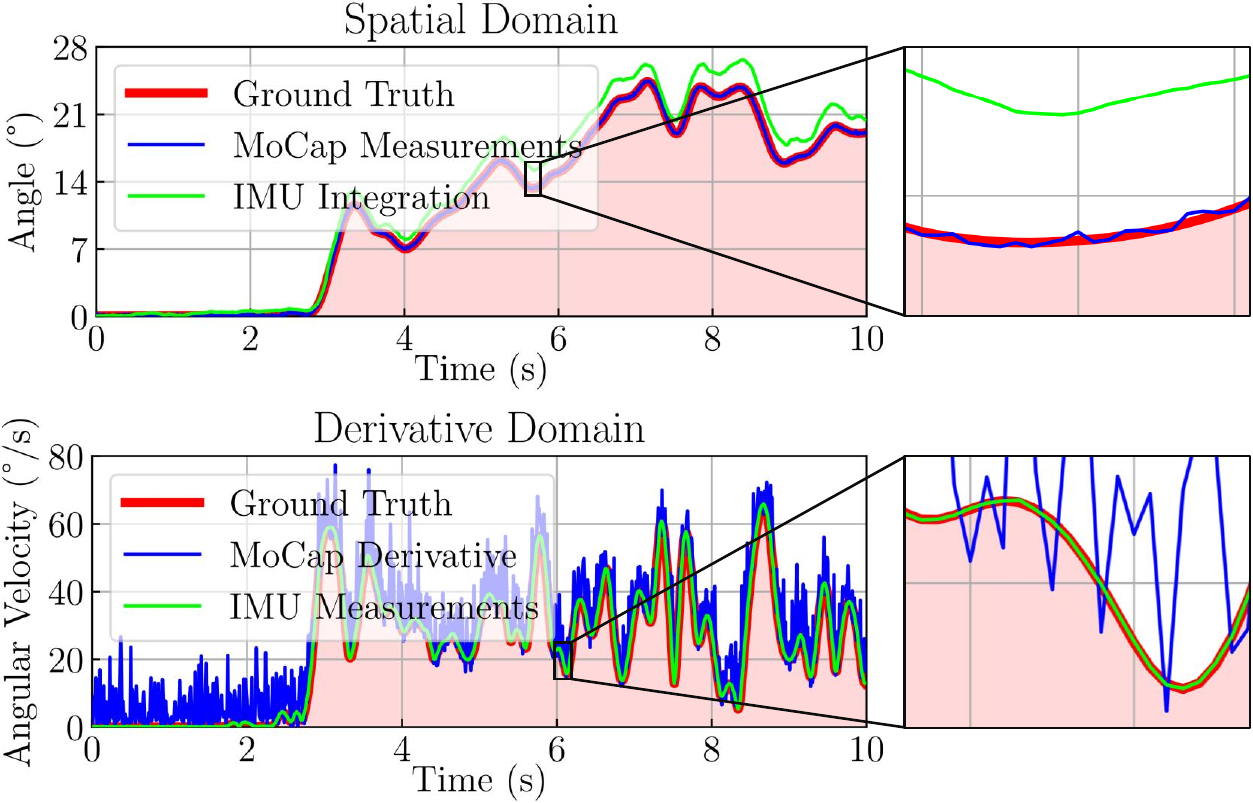}
\caption{Illustration of the complementary characteristics of MoCap and IMU data. The MoCap (blue) provides globally consistent measurements but introduces high-frequency noise, leading to significant errors in the derivative domain. Meanwhile, the IMU (green) offers robust derivative measurements but suffers from drift due to temporal integration.}
\label{FIG: Complementary characteristics}
\end{figure}

The differentiability of B-splines facilitates the construction of residuals in the derivative domain directly from IMU data. Crucially, incorporating additional IMU constraints is essential for our estimator because it enables the system to leverage the complementary strengths of inertial and MoCap data for error compensation. This mechanism is illustrated in \cref{FIG: Complementary characteristics} through simulation. Although MoCap provides globally consistent measurements with minimal errors, its high-frequency jitter is significantly amplified in the derivative domain. For example, a deviation of 0.1° in a single 100 Hz MoCap measurement corresponds to an angular velocity of 10°/s. In contrast, the IMU provides robust derivative measurements that inherently suppress high-frequency errors but experience cumulative spatial drift due to temporal integration. By combining the advantages of both modalities, the MLE approach can be employed to derive a high-precision motion trajectory.

\subsection{Batch Estimation}

\begin{figure}[t]
\centering
\includegraphics[width=\linewidth]{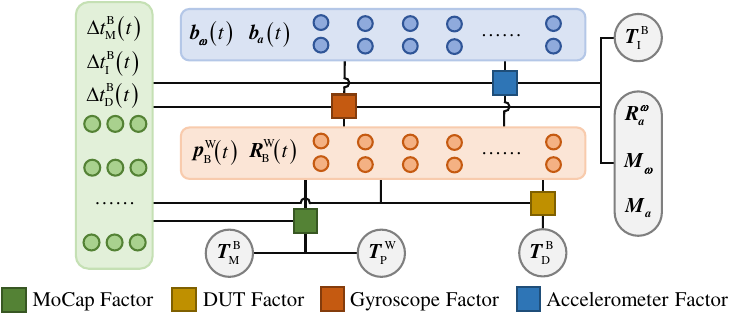}
\caption{Factor graph representation of the batch estimation problem, illustrating the connectivity between state variables and measurement factors from multiple sensors.}
\label{FIG: Factor graph}
\vspace{-10pt}
\end{figure}

The batch estimation objective is defined by integrating all the measurement factors. As illustrated in \cref{FIG: Factor graph}, the corresponding factor graph visually demonstrates how each residual is constructed from the state variables listed in \cref{TAB: Estimated states and their definitions}. Specifically, the problem is formulated as the minimization of the following negative log-likelihood function:
\begin{equation}
\begin{aligned}
\underset{\mathcal{X}}{\mathrm{argmin}} 
& \Biggl(  \sum_{i \in \mathcal{M} } 
\left \| \boldsymbol{r}_{\scriptscriptstyle\mathrm{M}} \left ( 
\tilde{\boldsymbol{T}}^{\scriptscriptstyle\mathrm{P}}_{\scriptscriptstyle\mathrm{M}_{i}}, \mathcal{X} \right ) \right \|_
{\boldsymbol{\Sigma}_{\mathrm{M} } }^ {2} + 
\sum_{j \in \mathcal{D} } 
\left \| \boldsymbol{r}_{\scriptscriptstyle\mathrm{D}} 
\left ( \tilde{\boldsymbol{T}}^{\scriptscriptstyle\mathrm{H}}_{\scriptscriptstyle\mathrm{D}_{i}}, 
\tilde{\boldsymbol{T}}^{\scriptscriptstyle\mathrm{H}}_{\scriptscriptstyle\mathrm{D}_{j}}, 
\mathcal{X} \right ) \right \|_
{\boldsymbol{\Sigma} _{\mathrm{D} } }^ {2} \\
& + \sum_{k \in \mathcal{I} } 
\left \| \boldsymbol{r}_{\scriptscriptstyle\mathrm{I,\ \boldsymbol\omega}}
\left ( \tilde{\boldsymbol{\omega}}_{\scriptscriptstyle\mathrm{I}_{i}}, 
\mathcal{X} \right ) \right \|_
{\boldsymbol{\Sigma} _{\mathrm{I}, \omega}}^ {2} + 
\sum_{k \in \mathcal{I} } 
\left \| \boldsymbol{r}_{\scriptscriptstyle\mathrm{I,\ \boldsymbol{a}}}
\left ( \tilde{\boldsymbol{a}}_{\scriptscriptstyle\mathrm{I}_{i}}, 
\mathcal{X} \right ) \right \|_
{\boldsymbol{\Sigma} _{\mathrm{I}, \boldsymbol{a}}}^ {2} \\
& + \sum_{k \in \mathcal{I} } 
\left \| \dot{\boldsymbol{b}}_{\scriptscriptstyle\boldsymbol{\omega}} \left( t \right) \right \|_
{\boldsymbol{\Sigma} _{\mathrm{I}, \boldsymbol{b}_{\scriptscriptstyle\boldsymbol{\omega}}}}^ {2} +
\sum_{k \in \mathcal{I} } 
\left \| \dot{\boldsymbol{b}}_{\scriptscriptstyle\boldsymbol{a}} \left( t \right) \right \|_
{\boldsymbol{\Sigma} _{\mathrm{I}, \boldsymbol{b}_{\scriptscriptstyle\boldsymbol{a}}}}^ {2}
\Biggr),
\end{aligned}
\end{equation}
where $\mathcal{M}$, $\mathcal{D}$, and $\mathcal{I}$ represent the measurements from the MoCap system, DUT, and auxiliary IMU, respectively. An additional residual term is introduced for the IMU bias, which follows a Wiener process driven by zero-mean white Gaussian noise. The covariance matrices $\boldsymbol{\Sigma} _{\mathrm{I}, \omega}$ and $\boldsymbol{\Sigma} _{\mathrm{I}, \boldsymbol{a}}$ for IMU readings, along with the random walk noise parameters $\boldsymbol{\Sigma} _{\mathrm{I}, \boldsymbol{b}_{\scriptscriptstyle\boldsymbol{\omega}}}$ and $\boldsymbol{\Sigma} _{\mathrm{I}, \boldsymbol{b}_{\scriptscriptstyle\boldsymbol{a}}}$, are determined through prior static calibration \cite{rehder2016extending}. The noise in MoCap data is assumed to be white Gaussian noise, with $\boldsymbol{\Sigma}_{\mathrm{M}}$ estimated using a similar calibration procedure. The covariance $\boldsymbol{\Sigma} _{\mathrm{D}}$ associated with DUT measurements is derived from \cref{SEC: Pose Factor}. Based on the initial guesses from \cref{SEC: Initialization}, this nonlinear least-squares problem is solved using the Ceres solver \cite{Agarwal_Ceres_Solver_2022}, ensuring optimal estimation by balancing errors across multiple data sources.

\section{Experiments}

In this section, we conduct a series of experiments to validate the proposed estimator, including: $\left( i \right)$ Simulation tests, which assess the accuracy of the estimated trajectory under controlled conditions. $\left( ii \right)$ Real-world experiments on both the self-collected dataset and the TUM-VI dataset, evaluating calibration precision and trajectory estimation accuracy. $\left( iii \right)$ Application to the practical SLAM benchmark tasks in XR, and discussion on the impact of GT accuracy on error metrics.

For comparison, we introduce four representative MoCap-based GT trajectory estimation methods: $\left( i \right)$ \textbf{Raw-HEC} \cite{jinyu2019survey}, a two-stage hand-eye calibration method that uses raw camera and IMU data from the DUT, calibrated against MoCap poses. $\left( ii \right)$ \textbf{P2P-HEC} \cite{furrer2018evaluation, shu2024spatiotemporal}, a pose-to-pose hand-eye calibration method that directly aligns the output poses of the DUT with the MoCap measurements. $\left( iii \right)$ \textbf{Vicon2GT} \cite{geneva2020vicon2gt}, which adopts a discrete-time state representation and performs batch estimation by combining IMU integration and absolute pose constraints. This approach is conceptually similar to the GT generation procedure used in the EuRoC dataset \cite{burri2016euroc}. $\left( iv \right)$ \textbf{Kalibr-M} \cite{rehder2016extending}, the MoCap extension of Kalibr, which performs continuous-time MLE based on IMU data and absolute pose measurements. To assess the effect of modeling time offset as a time-varying state, we evaluate two variants of our proposed high-precision GT estimator (\textbf{HPGT}): a version with a fixed time offset (\textbf{HPGT-Fix}) and the full method incorporating variable time offset (\textbf{HPGT-Var}). Since Raw-HEC and P2P-HEC only perform spatiotemporal calibration, we apply a median filter as done in the TUM-VI dataset \cite{schubert2018tum} to mitigate jitter caused by the MoCap system. Vicon2GT, Kalibr-M, and our method compensate for jitter using the auxiliary IMU in our setup, which consequently requires additional spatiotemporal calibration. 

In the following comparative experiments, the self-collected dataset was acquired using the Vicon Vero 2.2 MoCap system and the auxiliary IMU HiPNUC-14R5. The sensor calibration parameters are listed in \cref{TAB: Noise parameters used in our simulator}, where the noise levels of the auxiliary IMU are more than half an order of magnitude lower than those of consumer-grade IMUs commonly used in XR applications.

\subsection{Simulation Study}

\begin{table}[t]
\caption{Discrete-time noise parameters of the 300 Hz MoCap system and 500 Hz auxiliary IMU, used in both our experimental setup and simulator}
\label{TAB: Noise parameters used in our simulator}
\centering
\setlength{\tabcolsep}{0.3cm}
\renewcommand{\arraystretch}{0.9}
\scriptsize
\begin{tabular}{l|c}
\toprule[1.0pt]
\multicolumn{1}{c|}{Noise Type} & Value \\[0.2ex] 
\midrule
MoCap translational noise density (m) & $4.2 \times 10^{-4}$ \\
MoCap rotational noise density (°) & $1.0 \times 10^{-1}$ \\
Accelerometer noise density (m/$\mathrm{s}^2$) & $1.1 \times 10^{-2}$ \\
Accelerometer random walk (m/$\mathrm{s}^2$) & $2.1 \times 10^{-6}$ \\
Gyroscope noise density (°/s) & $4.8 \times 10^{-2}$ \\
Gyroscope random walk (°/s) & $1.4 \times 10^{-5}$ \\
Time offset drift (ms/min) & $1.0$ \\
\bottomrule[1.0pt]
\end{tabular}
\end{table}

The estimator is initially validated through simulation experiments where the GT trajectory is directly available, enabling precise error analysis with minimal uncertainty. The simulator is implemented based on Blender \cite{Blender}, employing a virtual indoor scene from the Replica dataset \cite{straub2019replica}, with necessary environmental settings. A predefined trajectory is used to generate MoCap and IMU data, with random spatiotemporal extrinsic parameters applied. Following standard practice, IMU readings are corrupted by white noise and random walk bias, while MoCap measurements are subject to additive white noise. Taking the IMU clock as the reference, the time offsets of other sensors are modeled as time-varying variables scaled by a fixed factor. The noise standard deviations (STDs) are set to match those of the devices used in our real-world experiments, as listed in \cref{TAB: Noise parameters used in our simulator}. Additionally, stereo images are rendered based on the motion trajectory in Blender, which are input to ORB-SLAM3 \cite{campos2021orb} to obtain the trajectory, serving as the DUT data.

\begin{figure}[t]
\centering
\includegraphics[width=0.95\linewidth]{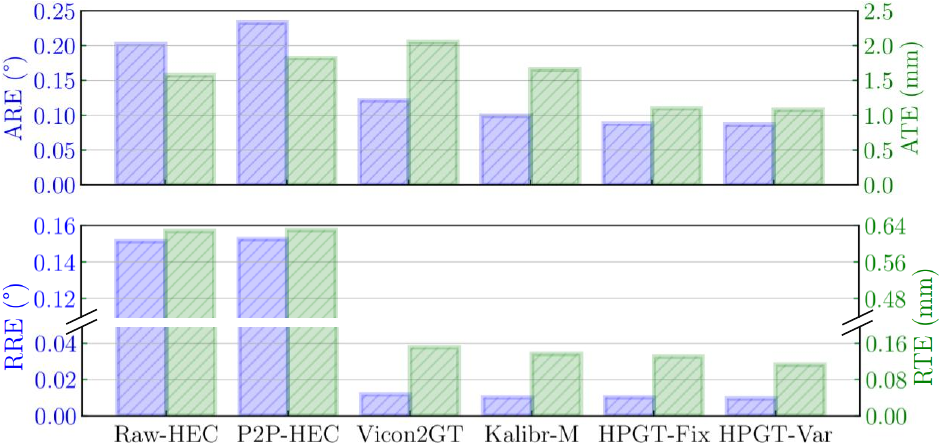}
\caption{Accuracy comparison of different methods on the simulation dataset, reporting absolute error (ARE/ATE) and inter-frame relative error (RRE/RTE) of the estimated trajectories.}
\label{FIG: Accuracy comparison on simulator}
\vspace{-10pt}
\end{figure}

Different methods were used to estimate trajectories expressed in the DUT frame, unified to 90 Hz. The accuracy of each algorithm was evaluated by comparing the estimated trajectories with the GT from the simulator. Root mean square error (RMSE) was computed to quantify trajectory errors, as shown in \cref{FIG: Accuracy comparison on simulator}. One of the main sources of absolute error is the extrinsic calibration. Since Vicon2GT and Kalibr-M struggle with accumulated errors in DUT trajectories, they exhibit significant inaccuracies. In contrast, our method effectively mitigates this issue, achieving a precise extrinsic calibration. Introducing a variable time offset improves the calibration accuracy, but the improvement is limited. MoCap jitter also contributes to absolute errors. Simple median filtering offers only limited compensation, leading to increased errors in Raw-HEC and P2P-HEC. On the other hand, inter-frame relative errors are primarily caused by MoCap jitter. As a result, Raw-HEC and P2P-HEC exhibit significantly higher relative errors. In contrast, the other methods incorporate auxiliary IMU measurements and are more effective in handling this high-frequency error. Overall, among the compared methods, HPGT-Var achieves the best performance, yielding the lowest absolute and relative errors.

\begin{table*}[!htp]
\caption{Comparison of calibration results among different methods on the self-collected and TUM-VI datasets. Mean and STD over multiple calibration trials are reported, with the best results highlighted in \textbf{bold}.}
\label{TAB: Calibration results}
\centering
\scriptsize
\begin{tabular}{c@{\hskip 4pt}|c@{\hskip 0.8pt}cc@{\hskip 9pt}c|c@{\hskip 0.8pt}cc@{\hskip 9pt}cc}
% \begin{tabular}{c|cccc|ccccc}
\toprule[1.0pt]
\multirow{3}{*}{\rule{0pt}{4ex}Method} & \multicolumn{4}{c|}{Self-Collected Dataset} & \multicolumn{4}{c}{TUM-VI Dataset} \\ 
\addlinespace[-0.4pt] \cmidrule{2-9} \addlinespace[-0.8pt]
 & \multicolumn{2}{c|}{MoCap-IMU} & \multicolumn{2}{c|}{MoCap-DUT} & \multicolumn{2}{c|}{MoCap-IMU} & \multicolumn{2}{c}{MoCap-DUT} \\
\addlinespace[-0.4pt] \cmidrule{2-9} \addlinespace[-0.8pt]
 & $\theta$ (°) & \multicolumn{1}{c|}{$p$ (cm)} & $\theta$ (°) & $p$ (cm) & $\theta$ (°) & \multicolumn{1}{c|}{$p$ (cm)} & $\theta$ (°) & $p$ (cm) \\
\addlinespace[-0.4pt] \midrule \addlinespace[1.4pt]
Raw-HEC & / & \multicolumn{1}{c|}{/} & 114.042±0.078 & 2.467±0.102 & / & \multicolumn{1}{c|}{/} & 1.284±0.072 & 7.223±\textbf{0.081} \\
P2P-HEC & / & \multicolumn{1}{c|}{/} & 113.891±0.095 & 2.493±0.129 & / & \multicolumn{1}{c|}{/} & 1.328±0.077 & 7.377±0.112 \\
Vicon2GT & 27.093±0.082 & \multicolumn{1}{c|}{6.912±0.103} & 114.067±0.116 & 2.407±0.176 & 1.324±0.089 & \multicolumn{1}{c|}{7.225±0.122} & 1.301±0.086 & 7.151±0.131 \\
Kalibr-M & 27.089±0.051 & \multicolumn{1}{c|}{6.764±0.083} & 114.031±0.108 & 2.307±0.194 & 1.218±\textbf{0.061} & \multicolumn{1}{c|}{7.321±0.097} & 1.292±0.079 & 7.324±0.107 \\
HPGT-Fix (ours) & 27.132±0.044 & \multicolumn{1}{c|}{6.833±\textbf{0.066}} & 113.936±0.068 & 2.510±0.110 & 1.269±0.066 & \multicolumn{1}{c|}{7.288±0.081} & 1.266±0.070 & 7.249±0.083 \\
HPGT-Var (ours) & 27.121±\textbf{0.041} & \multicolumn{1}{c|}{6.842±0.067} & 113.921±\textbf{0.062} & 2.481±\textbf{0.094} & 1.274±0.063 & \multicolumn{1}{c|}{7.271±\textbf{0.078}}  & 1.271±\textbf{0.068} & 7.264±0.085 \\ 
\bottomrule[1.0pt]
\end{tabular}
\end{table*}

\subsection{Real-World Experiment}

\subsubsection{Extrinsic Calibration}

In addition to simulation tests, real-world experiments were conducted on both self-collected and public datasets. First, extrinsic calibration performance was evaluated, as it greatly affects trajectory accuracy and enables direct comparison. Specifically, multiple calibrations were performed using the same set of devices, where higher consistency in the calibration results indicates better accuracy. As shown on the left side of \cref{FIG: Teaser figure}, a SLAM benchmarking system was constructed, incorporating a MoCap system, an auxiliary IMU, and our XR prototype (DUT). Ten data sequences were collected to compare different algorithms. Furthermore, validation was conducted on four IMU calibration sequences from the TUM-VI dataset, which contains MoCap, IMU, and camera data. ORB-SLAM3 \cite{campos2021orb} was employed to estimate the DUT trajectory using IMU and camera measurements.

\begin{figure}[t]
\centering
\includegraphics[width=\linewidth]{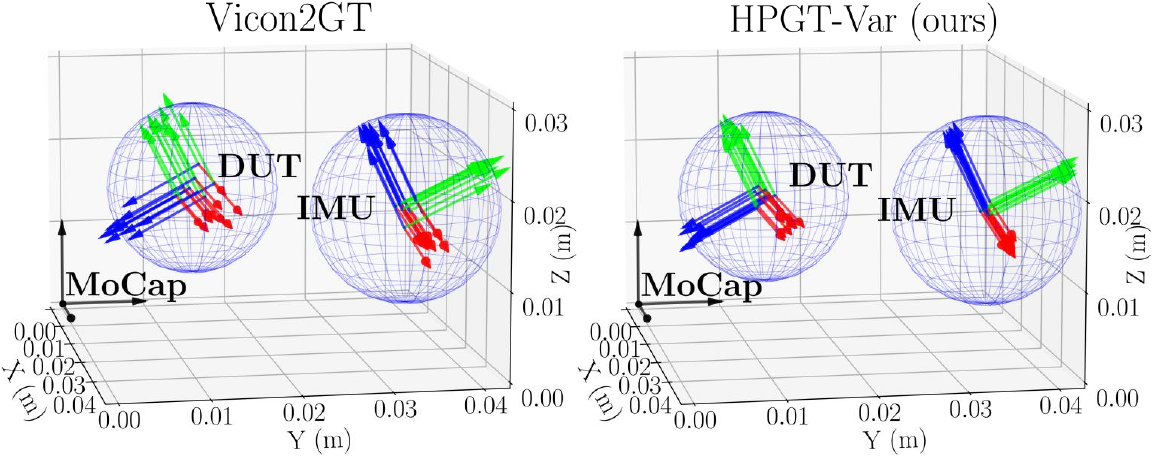}
\caption{Visualization of calibration repeatability. A 10 mm radius sphere centered at the mean position of the calibration results is plotted as a reference. A higher concentration near the center of the sphere indicates greater accuracy.}
\label{FIG: Visualization of calibration repeatability}
\vspace{-10pt}
\end{figure}

Different methods were used to estimate the extrinsic parameters between MoCap and both the IMU and DUT. To facilitate comparison, the extrinsics are represented with 2-DoF: the rotation angle $\theta$ and the translation magnitude $p$, with the mean and STD reported. The comparison analysis is presented in \cref{TAB: Calibration results}. Calibration results show some discrepancies among different algorithms, with mean differences reaching 0.18°/2.03 mm on the self-collected dataset and 0.11°/2.26 mm on the TUM-VI dataset. Regarding STD, HPGT-Var achieves the best performance, particularly for MoCap-DUT calibration, where it remains below 0.07° and 0.94 mm. This advantage is attributed to our dedicated DUT residual modeling and variable time offset estimation. Notably, although Raw-HEC leverages DUT raw images and additional calibration board information to avoid accumulated trajectory errors, our method still achieves competitive calibration accuracy. Moreover, since our approach does not rely on raw sensor data, it offers greater generality and can be applied to commercial devices that only provide pose outputs. To visually illustrate the calibration consistency across 6-DoF, we show the spatial distribution of the results for Vicon2GT and HPGT-Var on the self-collected dataset in \cref{FIG: Visualization of calibration repeatability}. Using the MoCap body frame as the reference, we plot the frames of the DUT and IMU after each process, with HPGT-Var demonstrating higher repeatability.

\subsubsection{Jitter Mitigation}

In most SLAM research, inter-frame jitter in raw MoCap data is largely overlooked. It is thus important to demonstrate its presence and emphasize its impact on SLAM benchmarking for XR applications. Quantitative analysis of jitter on real-world datasets is challenging due to the absence of GT, particularly in public datasets. Therefore, we begin with a qualitative analysis and concurrently demonstrate the effectiveness of our method in mitigating jitter. As shown on the right side of \cref{FIG: Teaser figure}, we compare the raw MoCap data with the results estimated by HPGT-Var, and plot the DUT trajectory as an additional reference. To enhance clarity, the jitter in the raw MoCap data is amplified threefold relative to the estimated trajectory. High-frequency jitter is observed in both our self-collected and TUM-VI datasets. After applying our estimator, the trajectory becomes smooth, stable and correctly aligned with the DUT frame.

\begin{figure}[t]
\centering
\includegraphics[width=0.92\linewidth]{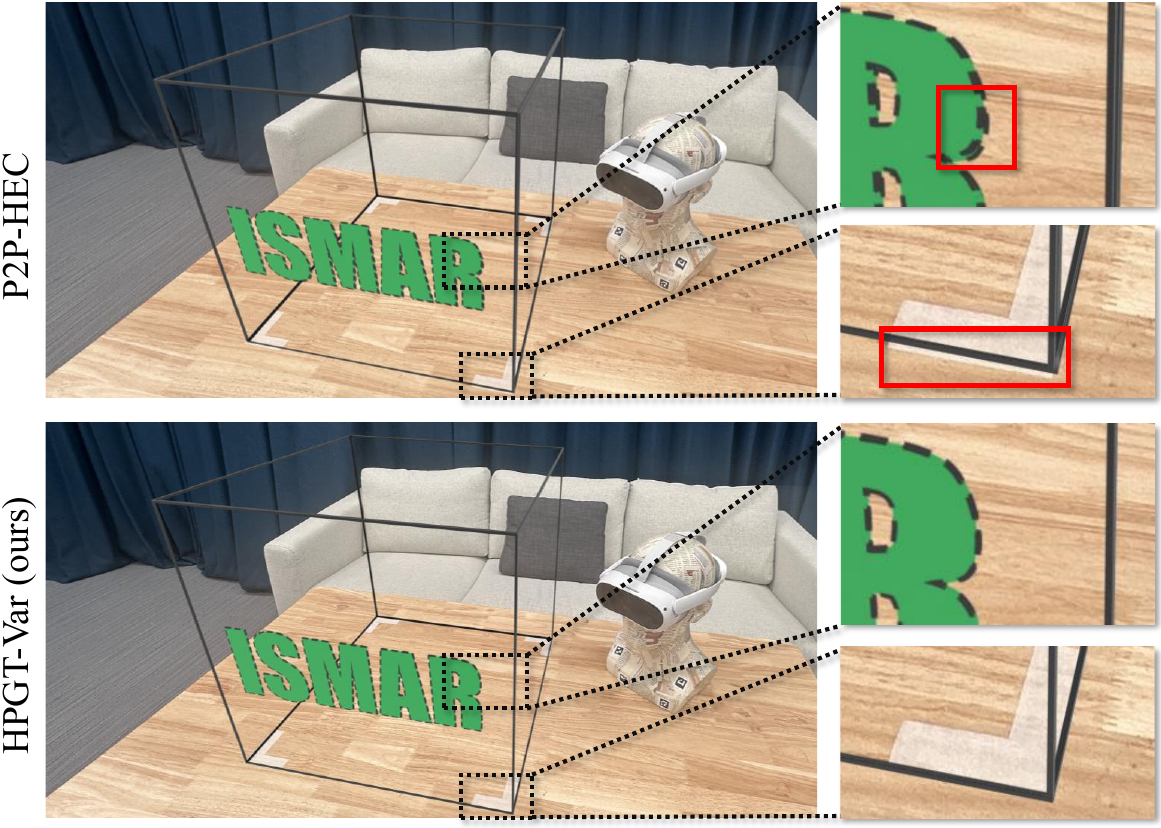}
\caption{AR rendering using trajectories estimated by P2P-HEC and our HPGT-Var. The results are rendered from a fixed viewpoint, with virtual objects overlaid across three consecutive frames. The zoomed-in view highlights the effects of trajectory errors.}
\label{FIG: AR rendering}
\end{figure}

Based on our XR prototype, we developed an AR application to further demonstrate the effect of jitter elimination. After calibration, virtual content can be rendered on the DUT's video see-through camera using trajectories estimated by different methods to reflect their accuracy. We first placed a reference marker in the real world and aligned a virtual cube with them from a specific camera pose. The DUT was then moved and held steady at a different viewpoint. At this point, virtual objects were rendered using trajectories estimated by P2P-HEC and HPGT-Var, and their results over three consecutive frames were overlaid. In the absence of noise, the virtual object should remain static and aligned with the real scene. However, due to the inability of P2P-HEC to handle inter-frame errors, noticeable artifacts appear in the overlaid rendering, as shown in \cref{FIG: AR rendering}. This indicates that the trajectory estimated by P2P-HEC fails to achieve the required accuracy for benchmarking SLAM jitter in the XR field, which may lead to inaccurate metrics. In contrast, HPGT-Var effectively mitigates this problem by compensating for IMU errors, resulting in stable and accurate rendering. Additionally, the result obtained using P2P-HEC showed a misalignment between the virtual object and reference markers, which may be due to calibration errors.

\subsubsection{Comprehensive Evaluation}

\begin{figure}[t]
\centering
\includegraphics[width=0.98\linewidth]{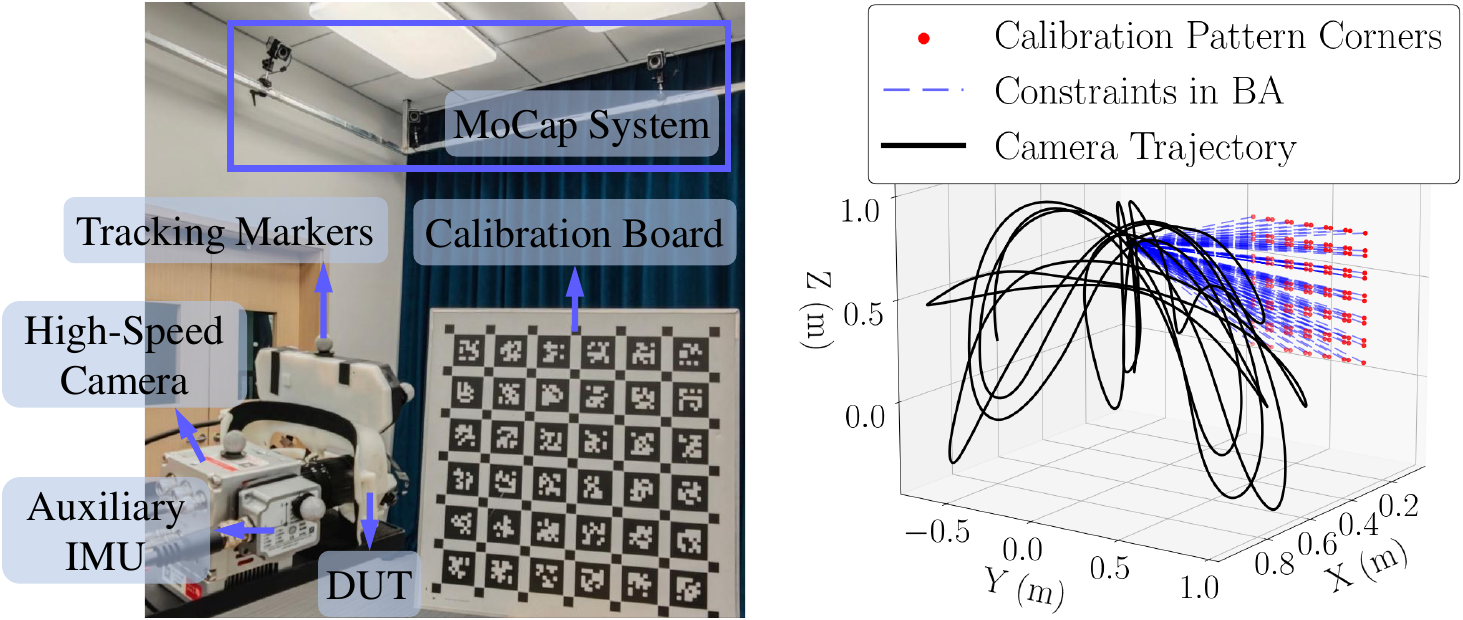}
\caption{Experimental setup for quantitative evaluation of trajectory accuracy (left) and an illustration of trajectory reconstruction using the BA algorithm (right).}
\label{FIG: Experimental setup}
\vspace{-10pt}
\end{figure}

\begin{table}[]
\caption{Comparison of trajectory estimation accuracy of different methods on self-collected datasets. Absolute errors (ARE/ATE) and inter-frame relative errors (RRE/RTE) are reported. Results meeting the target metrics$^*$ are highlighted in \colorbox{green!30}{green}, with the best results shown in \textbf{bold}.}
\label{TAB: Trajectory estimation accuracy}
\centering
\scriptsize
% \begin{tabular}{c|c|cccc}
\begin{tabular}{c@{\hskip 2mm}|c@{\hskip 2mm}|c@{\hskip 1mm}c@{\hskip 1mm}c@{\hskip 1mm}c@{\hskip 1mm}}
\toprule[1.0pt]
Scenario             & Algorithm 
& \begin{tabular}[c]{@{}c@{}}ARE\\ (°)\end{tabular} 
& \begin{tabular}[c]{@{}c@{}}ATE\\ (mm)\end{tabular} 
& \begin{tabular}[c]{@{}c@{}}RRE\\ (°)\end{tabular} 
& \begin{tabular}[c]{@{}c@{}}RTE\\ (mm)\end{tabular} \\ 
\midrule
\multirow{6}{*}{\begin{tabular}[c]{@{}c@{}} \rule{0pt}{5.2ex} Sufficient \\ Motion \end{tabular}} 
& Raw-HEC & 0.269   & \colorbox{green!30}{1.831}    & 0.145   & 0.520    \\
& \rule{0pt}{8pt} P2P-HEC & 0.294   & 2.332    & 0.149   & 0.521    \\
& Vicon2GT   & 0.273   & 3.188    & \colorbox{green!30}{0.017}   & \colorbox{green!30}{0.172}    \\
& Kalibr-M   & \colorbox{green!30}{0.184}   & 2.041    & \colorbox{green!30}{0.014}   & \colorbox{green!30}{0.156}    \\
& HPGT-Fix (ours)   & \colorbox{green!30}{0.164}   & \colorbox{green!30}{1.365}    & \colorbox{green!30}{0.014}   & \colorbox{green!30}{0.142}    \\
& HPGT-Var (ours)   & \colorbox{green!30}{\textbf{0.151}}   & \colorbox{green!30}{\textbf{1.341}}    & \colorbox{green!30}{\textbf{0.012}}   & \colorbox{green!30}{\textbf{0.133}}    \\ 
\midrule
\multirow{6}{*}{\begin{tabular}[c]{@{}c@{}} \rule{0pt}{5.2ex} Motion \\ Degradation \end{tabular}} 
& Raw-HEC & 0.276   & 2.105    & 0.141   & 0.526    \\
& \rule{0pt}{8pt} P2P-HEC & 0.315   & 2.768    & 0.142   & 0.528    \\
& Vicon2GT   & 0.326   & 3.971    & \colorbox{green!30}{0.018}   & \colorbox{green!30}{0.175}    \\
& Kalibr-M   & 0.202   & 2.817    & \colorbox{green!30}{0.014}   & \colorbox{green!30}{0.163}    \\
& HPGT-Fix (ours)   & \colorbox{green!30}{0.169}   & \colorbox{green!30}{\textbf{1.792}}    & \colorbox{green!30}{\textbf{0.013}}   & \colorbox{green!30}{0.152}    \\
& HPGT-Var (ours)   & \colorbox{green!30}{\textbf{0.162}}   & \colorbox{green!30}{1.811}    & \colorbox{green!30}{\textbf{0.013}}   & \colorbox{green!30}{\textbf{0.143}}    \\ 
\bottomrule[1.0pt]
\multicolumn{6}{@{}l@{}}{\small{$^*$Target outlined in \cref{SEC: Introduction}: ARE/ATE $<$ 0.2°/2 mm, RRE/RTE $<$ 0.02°/0.2 mm.}}
\end{tabular}
\vspace{-8pt}
\end{table}

\begin{table*}[t]
\scriptsize
\caption{SLAM benchmark results of leading XR devices and open-source algorithms. Results using our HPGT-Var as GT are shown in \textbf{bold}, and those from the P2P-HEC are provided in parentheses for comparison.}
\label{TAB: SLAM benchmark result}
\setlength{\tabcolsep}{5pt}
\centering
\scriptsize
\begin{tabular}{c|cccc|cccc}
\toprule[1.0pt]
\multirow{2}{*}{\rule{0pt}{4ex} DUT} & \multicolumn{4}{c|}{Standard Scenario} & \multicolumn{4}{c}{Challenging Scenarios} \\
\cmidrule{2-9}
                     & ARE (°)      & ATE (mm)     & RRE (°)      & RTE (mm)     & ARE (°)      & ATE (mm)     & RRE (°)      & RTE (mm)     \\
\midrule
AVP                  & \textbf{0.739} (0.921) & \textbf{10.117} (11.032) & \textbf{0.032} (0.153) & \textbf{0.397} (0.844) & \textbf{0.821} (1.045) & \textbf{12.381} (13.243) & \textbf{0.038} (0.161) & \textbf{0.408} (0.871) \\
Quest 3              & \textbf{0.696} (0.897) & \textbf{12.310} (13.193) & \textbf{0.049} (0.164) & \textbf{0.431} (0.912) & \textbf{0.794} (1.028) & \textbf{16.534} (17.432) & \textbf{0.059} (0.173) & \textbf{0.501} (0.932) \\
PICO 4               & \textbf{0.714} (0.949) & \textbf{13.415} (14.417) & \textbf{0.045} (0.167) & \textbf{0.435} (0.927) & \textbf{1.082} (1.397) & \textbf{15.965} (17.221) & \textbf{0.061} (0.185) & \textbf{0.486} (0.937) \\
ORB-SLAM3            & \textbf{0.821} (1.013) & \textbf{23.657} (24.714) & \textbf{0.059} (0.175) & \textbf{0.684} (1.105) & \textbf{1.211} (1.468) & \textbf{42.541} (44.317) & \textbf{0.070} (0.193) & \textbf{0.797} (1.213) \\
\bottomrule[1.0pt]
\end{tabular}
\vspace{-5pt}
\end{table*}

Ultimately, we aim to quantitatively evaluate the accuracy of the trajectory estimated by our method, which serves as a comprehensive indicator of the performance of both calibration and jitter mitigation. This evaluation is essential to verify whether our method meets SLAM benchmark requirements in XR applications. To this end, a higher-precision localization system is required to provide 6-DoF reference trajectories. We employ a high-speed camera with a resolution of 2K at 1000 Hz and a field of view (FOV) of 56° × 32°, and set up the experiment system as shown on the left side of \cref{FIG: Experimental setup}. The camera captures a calibration board at close range, and the trajectory is reconstructed via bundle adjustment (BA), as illustrated on the right side of \cref{FIG: Experimental setup}. Due to the short imaging distance and minimal motion blur, the BA algorithm achieves an average reprojection error below 0.1 pixel. Under the system configuration, this corresponds to a pose accuracy of approximately 0.05 mm in translation and 0.003° in rotation, which ensures sufficient precision for generating reference trajectories. It should be noted that this setup has a limitation: the motion range is restricted because the calibration board must remain within the FOV of the high-speed camera. As a result, this approach can only be used to verify the accuracy of our estimator within a confined space.

During the camera motion, data is simultaneously collected from the rigidly connected MoCap system, auxiliary IMU, and the DUT. We collected five sequences in each of the two scenarios for the comparative experiments. The first scenario involves sufficient motion excitation, while the second features motion degradation, which is primarily used to evaluate the robustness of the algorithm. Each sequence has a duration of approximately 1 minute. The output trajectory frequencies of the comparative algorithms are uniformly set to 90 Hz. In addition, the reference trajectory reconstructed by the high-speed camera is transformed into the DUT frame to facilitate evaluation. The trajectory accuracy quantification results of different methods are presented in \cref{TAB: Trajectory estimation accuracy}. Real-world experiments yield conclusions similar to those from simulation tests. As Raw-HEC and P2P-HEC do not effectively mitigate the jitter present in the raw MoCap data, they produce large absolute and relative errors, with P2P-HEC showing the highest overall error among all evaluated methods. Both Vicon2GT and Kalibr-M achieve small relative errors but suffer from inaccurate calibration with the DUT, leading to large absolute errors. In contrast, our method achieves high-precision localization trajectory estimation and is also robust in motion-degraded scenarios. Notably, only our algorithm meets all the accuracy requirements outlined in \cref{SEC: Introduction}.

\subsection{Application to SLAM Benchmarking}

\begin{figure}[t]
\centering
\includegraphics[width=0.92\linewidth]{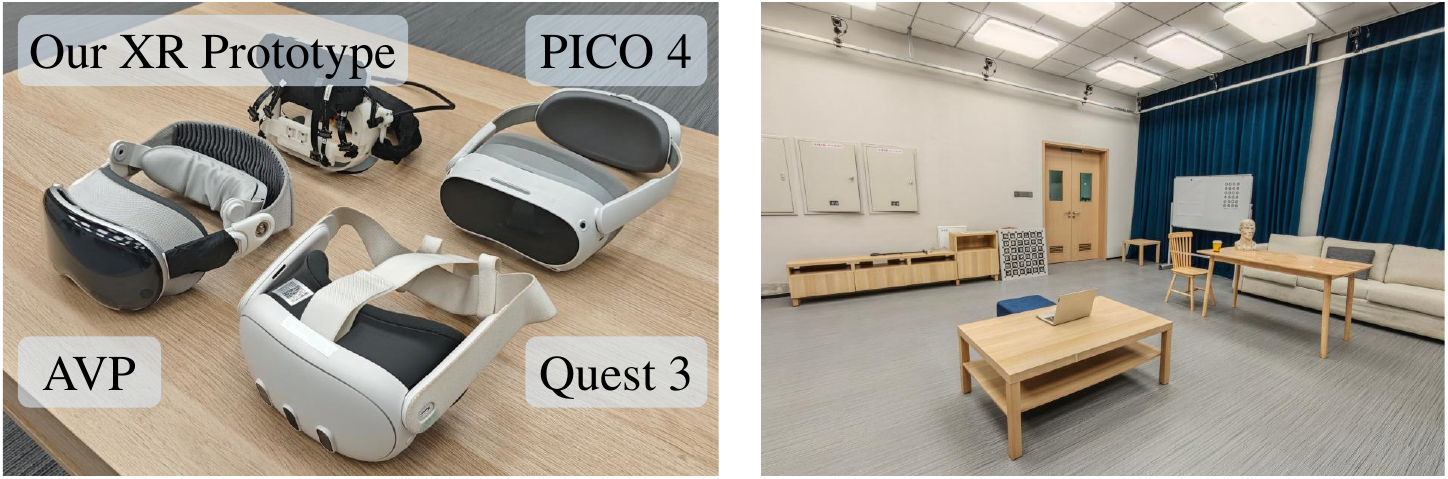}
\caption{DUTs for SLAM benchmarking (left) and the test environment (right). The DUTs include AVP, Quest 3, and PICO 4. Additionally, our XR prototype is used for collecting the SLAM sensor data as the inputs to open-source SLAM algorithms. }
\label{FIG: DUTs and scenario}
\vspace{-10pt}
\end{figure}

To demonstrate the practical value of our method and emphasize the importance of high-precision GT trajectories for SLAM benchmarking in XR, we apply it to real-world use cases. Specifically, we benchmark three commercial XR devices with SLAM capabilities, including Apple Vision Pro (AVP)\footnote{\fontsize{8pt}{10pt}\selectfont \url{https://developer.apple.com/visionos}; Apple visionOS V2.3.1}, Meta Quest 3\footnote{\fontsize{8pt}{10pt}\selectfont \url{https://developers.meta.com/horizon/develop}; Meta Horizon OS V74.1020}, and PICO 4\footnote{\fontsize{8pt}{10pt}\selectfont \url{https://developer.picoxr.com/resources}; PICO OS V5.12.0.S}. The trajectories from these commercial devices are obtained via manufacturer-provided software development kits (SDKs) that interface with their onboard SLAM algorithms within a Unity-based environment. We also benchmark the SOTA open-source SLAM algorithm ORB-SLAM3 \cite{campos2021orb}, with input data collected using our XR prototype. Experiments are conducted in our MoCap laboratory, where the environment is designed to mimic a typical indoor home setting, simulating real-world XR usage. The DUTs and the test environment are shown in \cref{FIG: DUTs and scenario}. In addition to the standard setting, we introduce a challenging scenario with low lighting and rapid motion. Each DUT performs five one-minute sequences per scenario, and the average error metrics are reported. In addition to the commonly used metric ATE, we also include ARE and inter-frame relative errors, which are critical for XR applications. For comparison, we compute error metrics with reference to the trajectories estimated by both the P2P-HEC and HPGT-Var. This highlights the limitations of conventional MoCap-based trajectory estimation methods in XR SLAM benchmarking. 

\begin{figure}[t]
\centering
\includegraphics[width=\linewidth]{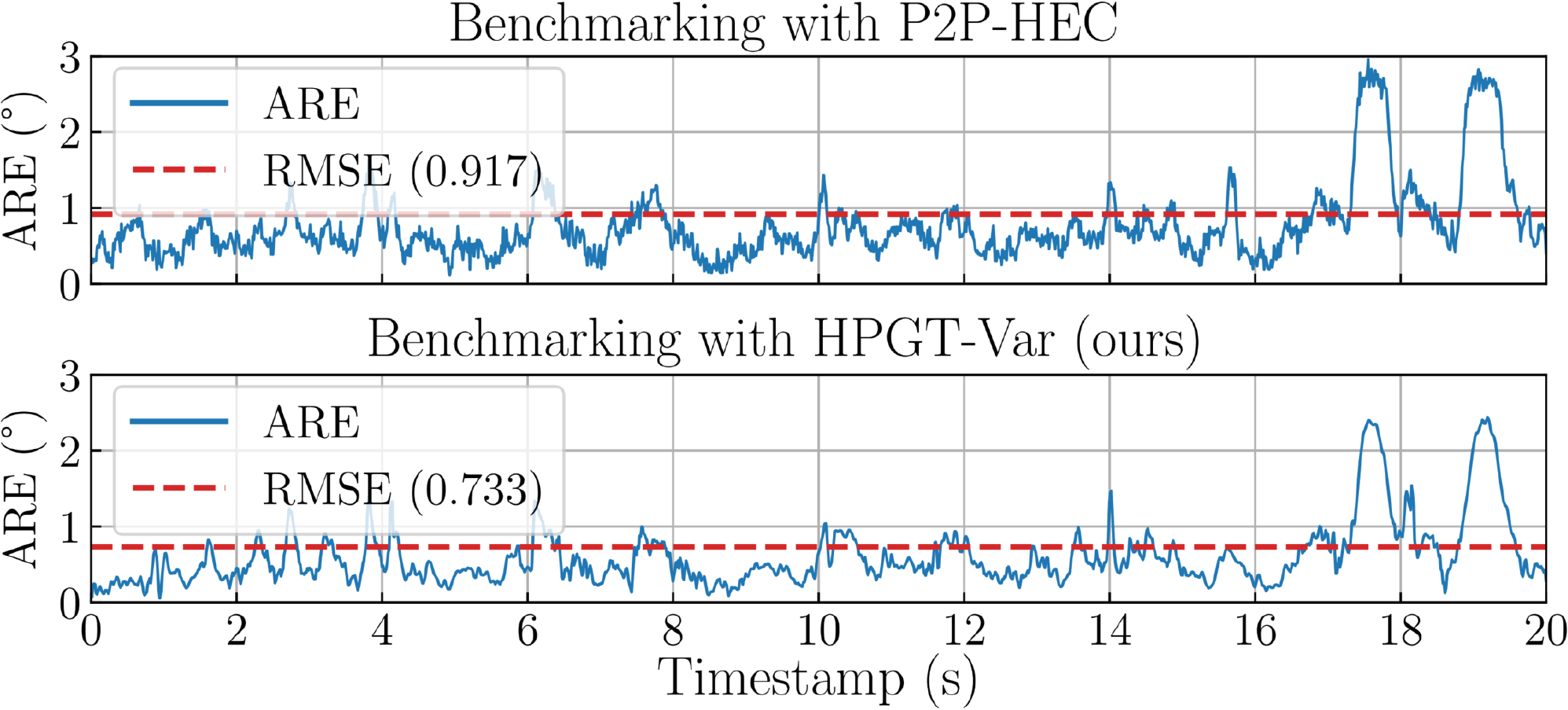}
\caption{Comparison of the ARE over time for the AVP, obtained using different trajectory estimation methods. A 20-second segment was selected for analysis.}
\label{FIG: Comparison of ARE for AVP}
\vspace{-10pt}
\end{figure}

The benchmarking results are shown in \cref{TAB: SLAM benchmark result}. Commercial SLAM systems performed well with comparable error magnitudes, while open-source algorithms showed larger errors, especially in translation. In the challenging scenario, all systems exhibited increased errors, particularly the PICO 4, whose absolute error was large enough to cause noticeable virtual object drift. Comparisons with the metrics derived from P2P-HEC (shown in parentheses) indicate that the accuracy of the GT trajectories has a substantial impact on benchmarking results. The effect is especially evident in the relative error metrics, where the results from P2P-HEC were largely dominated by noise. Regarding ARE, the results are also limited in precision due to the influence of calibration errors and jitter. To illustrate this, \cref{FIG: Comparison of ARE for AVP} shows a segment of ARE over time for AVP in scenario $\mathcal{V}_{\scriptscriptstyle{1}}$. While the overall trends from P2P-HEC and HPGT-Var are generally consistent, the metric from P2P-HEC shows systematic bias due to calibration errors and exhibit pronounced jitter. Crucially, since many SLAM benchmarking methods rely on trajectory estimation methods similar to the P2P-HEC, which only performs calibration with limited accuracy, they struggle to provide precise SLAM error metrics for XR devices.

\section{Conclusion}

This paper proposes a novel localization trajectory estimator to provide GT for SLAM benchmarking in XR applications. Within a continuous-time MLE framework, the data from the MoCap, auxiliary IMU, and DUT are jointly integrated and optimized. The method effectively handles cumulative errors in DUT trajectories and variable time offsets between sensors, enabling precise spatiotemporal calibration. Moreover, by leveraging the complementary mechanism of MoCap and IMU measurements, it achieves accurate trajectory estimation by balancing their error characteristics. Extensive experiments on both simulation and real-world datasets demonstrate that our approach outperforms existing methods. It achieves trajectory errors of ARE/ATE $<$ 0.162°/1.811 mm and RRE/RTE $<$ 0.013°/0.143 mm, meeting the accuracy requirements outlined in \cref{SEC: Introduction}. The proposed method overcomes the limitations of conventional approaches and enables comprehensive and precise quantification of SLAM errors such as rotational drift and inter-frame jitter, paving the way for high-precision benchmarking and ongoing development of XR SLAM systems.

Nonetheless, our method has certain limitations. For instance, calibration accuracy may degrade under insufficient motion excitation. It can be mitigated by identifying such segments and down-weighting them during optimization. Moreover, to improve computational efficiency in long-duration trajectory estimation, non-uniform B-splines can be used to adaptively adjust control point density, representing a promising direction for future research.

%% if specified like this the section will be omitted in review mode
\acknowledgments{%
    This work was supported by Key R\&D Program of Zhejiang (No. 2024SSYS0047) and Zhejiang Leading Innovative and Entrepreneurial Teams (No. 2023R01010).
}

\bibliographystyle{abbrv-doi-hyperref}

\bibliography{template}

\end{document}